\documentclass{article}
\usepackage{arxiv}

\usepackage[]{hyperref}

\usepackage{graphicx}
\usepackage{amsopn}
\usepackage{amssymb}
\usepackage{amsmath}
\usepackage{algorithm}
\usepackage[noend]{algpseudocode}
\usepackage{array}
\usepackage{float}
\usepackage[caption=false,font=footnotesize]{subfig}
\DeclareMathOperator*{\argmax}{arg\,max}

\usepackage{colortbl}
\definecolor{myred}{rgb}{0.8,0,0}

\title{Interactively shaping robot behaviour with unlabeled human instructions}

\author{
Anis Najar \\
  Laboratoire de Neurosciences Cognitives Computationnelles (LNC2)\\
  INSERM U960\\
  Paris \\
  \texttt{anis.najar@ens.fr} \\
   \And
 Olivier Sigaud \\
  Institute for Intelligent Systems and Robotics, \\
  Sorbonne Universit\'{e}, CNRS UMR 7222\\
  Paris\\
  \texttt{olivier.sigaud@upmc.fr} \\
   \And
 Mohamed Chetouani \\
  Institute for Intelligent Systems and Robotics, \\
  Sorbonne Universit\'{e}, CNRS UMR 7222\\
  Paris\\
  \texttt{mohamed.chetouani@upmc.fr} \\
}

\begin{document}
\maketitle

\begin{abstract}

In this paper, we propose a framework that enables a human teacher to shape a robot behaviour by interactively providing it with unlabeled instructions. We ground the meaning of instruction signals in the task-learning process, and use them simultaneously for guiding the latter.
We implement our framework as a modular architecture, named TICS (Task-Instruction-Contingency-Shaping) that combines different information sources: a predefined reward function, human evaluative feedback and unlabeled instructions.
This approach provides a novel perspective for robotic task learning that lies between Reinforcement Learning and Supervised Learning paradigms.
We evaluate our framework both in simulation and with a real robot.
The experimental results demonstrate the effectiveness of our framework in accelerating the task-learning process and in reducing the number of required teaching signals.

\keywords{Interactive Machine Learning \and Human-Robot Interaction \and Shaping \and Reinforcement Learning  \and Unlabeled Instructions}
\end{abstract}

\section{Introduction}
\label{introduction}
Over the last few years, substantial progress has been made in both machine learning \cite{mnih_atari_2013,mnih_human-level_2015,silver_mastering_2016} and robotics \cite{feng_optimization_2014}.
However, applying machine learning methods to real-world robotic tasks still raises several challenges.
One important challenge is to reduce training time, as state-of-the-art machine learning algorithms still require millions of iterations for solving real-world problems \cite{mnih_atari_2013,mnih_human-level_2015,silver_mastering_2016}.
Two complementary approaches for task learning in Robotics are usually considered: autonomous learning and interactive learning.

Autonomous learning frameworks, such as Reinforcement Learning  \cite{kober_reinforcement_2013} or Evolutionary Approaches \cite{doncieux_evolutionary_2015}, rely on a predefined evaluation function that enables the robot to autonomously evaluate its performance on the task. 
The main advantage of this approach is the autonomy of the learning process. 
The evaluation function being integrated on board, the robot is able to optimize its behaviour without requiring help from a supervisor. 
However, when applied to real-world problems, this approach suffers from several limitations.
First, designing an appropriate evaluation function can be difficult in practice \cite{kober_reinforcement_2013}.
Second, autonomous learning is based on autonomous exploration which results in slow convergence of the learning process, and thus limits the feasibility of such approach in complex real-world problems. 
Moreover, autonomous exploration may lead to dangerous situations where the robot can damage itself or other objects, or even harm surrounding humans. 
Safety is an important issue that has to be considered when designing such autonomous learning systems \cite{garcia_comprehensive_2015}.

By contrast, interactive learning relies on human teaching signals for guiding the robot throughout the learning process \cite{chernova_robot_2014}.
Several types of teaching signals can be provided, such as demonstrations \cite{argall_survey_2009}, instructions \cite{pradyot_integrating_2012} and evaluative feedback \cite{knox_interactively_2009}.
Interactive learning methods overcome the limitations of autonomous learning by ensuring faster convergence rates and safer exploration.
However, they come at the cost of human burden during the teaching process, and the cost of predetermining the meaning of teaching signals \cite{vollmer_pragmatic_2016}. 
Encoding the meaning of a teaching signal requires engineering skills, which limits the usability of such methods for non expert users.
Also, predefined teaching signals constrain the users in the way they can interact with the robot.
So, a twofold challenge for interactive learning methods is to minimize their interaction load, and to provide more freedom to non expert users in choosing their own preferred signals for interacting with the robot.

In this paper, we propose a novel framework for robotic task learning that combines the benefits of both autonomous learning and interactive learning approaches.
First, we consider reinforcement learning with a predefined reward function for ensuring the autonomy of the learning process.
Second, we consider two types of human-provided teaching signals, evaluative feedback and instructions, for accelerating the learning process. 
Moreover, we relax the constraint of predetermining the meaning of instruction signals by making the robot incrementally interpret their meaning during the learning process.
Our main contribution is to show that instructions can effectively accelerate the learning process, even without predetermining their meaning. 

\begin{figure}
\begin{center}
\subfloat[The learning process under standard RL framework.]{\includegraphics[scale=0.4]{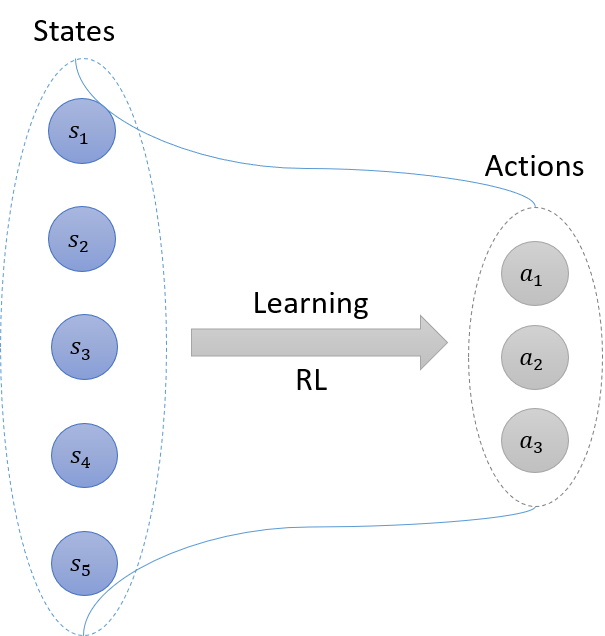}}
\hfill     
\subfloat[The learning process under our framework.]{\includegraphics[scale=0.4]{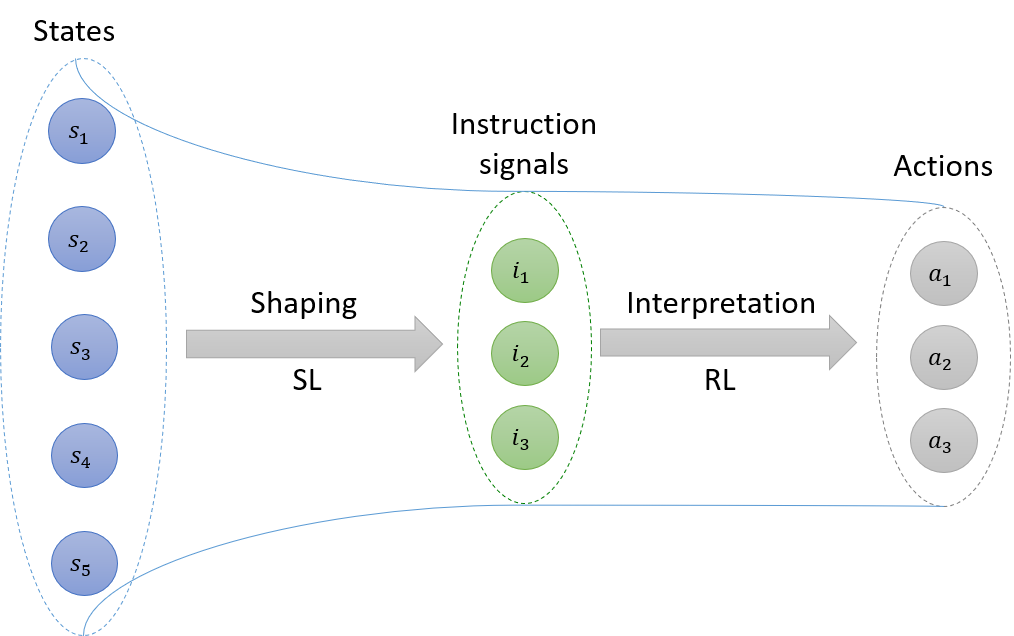}}
\caption{In our framework, unlabeled instructions are used for accelerating the learning process, by dividing it into two sub-problems: interpretation and shaping. Interpretation consists in mapping low-level instruction signals into actions and shaping consists in using the interpreted instructions for accelerating the learning process. Instructions are interpreted by reinforcement learning (RL) and used for shaping in a supervised learning (SL) way.}
\label{fig:sl-rm}
\end{center}
\end{figure}

We consider interactively provided instruction signals (\textit{e.g.} pointing to the left/to the right) that indicate to the robot which action it has to perform in a given situation (\textit{e.g.} turn left/turn right).
Our main idea is to use instruction signals as a means for transferring the information about the optimal action between several task states:
all states associated with the same instruction signal collectively contribute to interpreting the meaning of that signal; and in turn, an interpreted signal contributes to learning the optimal action in all task states to which it is associated. 
This scheme serves as a bootstrapping mechanism that reduces the complexity of the learning process; and constitutes a novel perspective for robotic task learning that lies between Reinforcement Learning and Supervised Learning paradigms. 
Under this scheme, unlabeled instructions (cf. definition in Section \ref{definitions}) are interpreted through a reinforcement learning process, and used for labeling task states in a supervised learning way (Fig. \ref{fig:sl-rm}).

We implement our framework as a modular architecture, named TICS (Task-Instruction-Contingency-Shaping), which combines different information sources: a predefined reward function, human evaluative feedback and unlabeled instructions.
We first evaluate our framework on two simulated tasks: object sorting and maze navigation.
Simulations allow us to systematically evaluate the performance of our system under different hypotheses about the teaching conditions, and to test its limits under worst case scenarios. 
For instance, we evaluate the robustness of our framework against various levels of sparse and erroneous teaching signals.
We then evaluate the framework on a real robotic platform.
The experimental results, obtained both in simulation and with the real robot, demonstrate the effectiveness of our framework in accelerating the task-learning process and in reducing the number of required teaching signals.

This paper is organized as follows. 
In the remainder of this section, we formulate our research question, provide some definitions, specify our assumptions and highlight the challenges we are facing.
In Section \ref{background}, we provide some background about robotic task learning.
We present our framework in Section \ref{model}.
The simulation protocol is presented in Section \ref{protocol}, and the results are reported in Section \ref{results}.
In Section \ref{baxter}, we evaluate our framework with a real robot.
The limitations of our work are discussed in Section \ref{discussion} before we conclude in Section \ref{conclusion}.

\section{Research question}

We consider a robot learning a task (\textit{e.g.} maze navigation) using an evaluation function such as a predefined reward function and/or evaluative feedback provided by a human teacher.  
In some situations, the teacher can provide the robot with an instruction that indicates which action it has to perform in the current situation (\textit{e.g.} turn left/turn right).
The robot is able to detect the low-level signals by which instructions are communicated (\textit{e.g.} pointing to the left/to the right). 
However, it does not know the meaning of each specific signal (\textit{i.e.} the action to which the signal refers).
The question we raise in this paper is how can the robot use these unlabeled instructions for learning the task?
We refer to this question as \textit{``interactively shaping robot behaviour with unlabeled human instructions"}.

\subsection{Definitions}
\label{definitions}
We now define the terms that we use in this paper:
\begin{itemize}

\item \textbf{Behaviour:} The sequence of actions that the robot performs in order to achieve a predefined task. 

\item \textbf{Shaping:} 
The mechanism by which a robot is influenced towards a desired behaviour. 

\item \textbf{Instruction:} Communicating an action to be performed in a given task state.

\item \textbf{Instruction signal:} The perceptual support through which an instruction is conveyed, for example a pointing gesture or a spoken word. In this work, we consider non-verbal instruction signals, namely human gestures.

\item \textbf{Meaning of an instruction signal:} The action to which the signal refers.

\item \textbf{Unlabeled instruction:} An instruction signal whose meaning is unknown to the robot.

\item \textbf{Evaluative feedback:} Communicating the correctness of a performed action. Also called critique. In this paper, we only consider binary feedback (correct/wrong). We also do not consider corrective feedback (instruction about past actions).

\end{itemize}

\subsection{Assumptions}
In this work, we consider the following assumptions: 
\begin{itemize}

\item \textbf{Elementary actions:} The robot is endowed with a set of predefined action primitives that are necessary for performing the task.

\item \textbf{Observable task states:} We consider a sequential task defined over a set of observable states. For each task state, there is an optimal action that has to be performed by the robot in order to complete the task.

\item \textbf{Evaluation function:} The robot has access to an evaluation function that allows it to find the optimal action for every task state. In this work, we consider a robot learning from a predefined reward function and/or from human evaluative feedback, using a reinforcement learning process.

\item \textbf{Discrete signals:} The robot is able to detect a predefined set of instruction signals.

\item \textbf{Unlabeled instructions:} In some states, the robot can receive an instruction signal. 
It knows that it is an instruction.
However, it does not know its meaning.

\end{itemize}

\subsection{Main challenges}
Our research question raises several challenges. 
First, we have different sources of information that need to be combined: the reward function, evaluative feedback and instructions.
These information sources are of different nature, so they may not be used computationally in exactly the same way. 
In addition, they can be in contradiction with each other. 
For instance, the information carried by the reward function can be different from the one communicated by the teacher. 
Consequently, these information sources must be combined properly.

Second, we need to take into account that the teacher is not perfect. 
For example, (s)he may not provide instructions for every situation or feedback for every performed action and (s)he can make mistakes and provide erroneous information. 
Consequently, we must evaluate the robustness of our solution against sparse and erroneous teaching signals.

Finally, as the interpretation of instructions is made interactively in the context of a task-learning process, we do not want to make the learning process longer than without using instructions, nor to put more burden on the teacher in using unlabeled instructions. 
Consequently, we must evaluate the cost of our method both in terms of convergence rate and interaction load.

\section{Background and Related work}
\label{background}

\subsection{Reinforcement Learning}

The Markov Decision Process (MDP) framework is a standard formalism for representing sequential decision-making problems \cite{sigaud_markov_2010}.
An MDP is defined as a tuple $<S,A,T,R,\gamma>$ where $S$ is the state-space and $A$ is an action-set.
$T: S \times A \to Pr(s'|s,a)$ defines a state-transition probability function, where $Pr(s'|s,a)$ represents the probability that the robot transitions from state $s$ to state $s'$ after executing action $a$.
$R: S \times A \to \mathbb{R}$ is a reward function that defines the reward $r(s,a)$ that the robot gets for performing action $a$ in state $s$.
When at time $t$, the robot performs an action $a_t$ from state $s_t$, it receives a reward $r_{t}$ and transitions to state $s_{t+1}$. 
The discount factor, $\gamma$, represents how much future rewards are taken into account for the current decision.

The behaviour of the robot is represented by a policy $\pi$ that defines a probability distribution over actions in every state $s \in S$: $\pi(s) = \{\pi(s,a);a \in A\} = \{Pr(a|s);a \in A\}$.
The quality of a policy is measured by the amount of rewards it enables the robot to collect over the long run.
The amount of cumulative rewards expected when starting from a state $s$ and following a policy $\pi$ is given by the state-value function and is written

\begin{equation} \label{eq:state-value}
V^{\pi}(s) = \sum\limits_a{\pi(s,a)[R(s,a)+ \gamma \sum\limits_{s'}}{Pr(s'|s,a) V^{\pi}(s')}].
\end{equation}

Another form of value function, called action-value function and noted $Q^{\pi}$, provides more directly exploitable information than $V^{\pi}$ for decision-making, as the agent has direct access to the value of each possible decision:

\begin{equation} \label{eq:action-value}
Q^{\pi}(s,a) = R(s,a)+ \gamma \sum\limits_{s'}{Pr(s'|s,a) V^{\pi}(s')} \quad; \forall s \in S, a \in A.
\end{equation}

To optimize its behaviour, the agent must find the optimal policy $\pi^*$ that maximizes $V^{\pi}$ and $Q^{\pi}$. 
The optimal policy can be derived using Reinforcement Learning algorithms such as Q-learning \cite{watkins_q-learning_1992} and Actor-Critic   \cite{barto_neuronlike_1983}.

The main idea of Actor-Critic architectures is to separately represent the value function (the critic) and the policy (the \textit{actor}).
The actor stores the policy $\pi$ and is used for action selection.
It is generally represented by a set of parameters $p(s,a)$ that reflect the preference for taking each action in each state.
At decision time $t$, the policy can be derived using a softmax distribution over the policy parameters: 
\[\pi_t(s,a) = Pr(a_t=a|s_t = s) = \frac{e^{p(s,a)}}{\sum_{b \in A} e^{p(s,b)}}.\]

The critic computes a value function that is used for evaluating the actions of the actor. The reward $r_{t}$ received at time $t$ is used for computing a temporal difference (TD) error

\begin{equation} \label{eq:dt}
\delta_t = r_{t} + \gamma V(s_{t+1}) - V(s_t).
\end{equation}

The TD error is then used for updating both the critic and the actor, using respectively Equations (\ref{eq:ac-vu}) and (\ref{eq:ac-pu}):

\begin{equation} \label{eq:ac-vu}
V(s_t) \leftarrow V(s_t) + \alpha \delta_t,
\end{equation}

\begin{equation} \label{eq:ac-pu}
p(s_t,a_t) \leftarrow p(s_t,a_t) + \beta \delta_t,
\end{equation}

where $\alpha$ and $\beta$ are two positive learning rates.
A positive TD error increases the probability of selecting $a_t$ in $s_t$, while a negative TD error decreases it. 

In Q-learning, the policy $\pi$ is not stored separately, but is derived from the Q-function at decision time, using the softmax distribution over the Q-values.
The Q-function is first initialized for every state-action pair. 
Then, it is iteratively updated after each transition using:

\begin{equation} \label{eq:q-learning}
Q(s_t,a_t) \gets  Q(s_t,a_t)+ \alpha [r_t + \gamma \max_{a' \in A} Q(s_{t+1},a') - Q(s_t,a_t)].
\end{equation}

\subsection{Interactive Learning}

\subsubsection{Shaping with evaluative feedback}
Delivering evaluative feedback is an intuitive way for training a robot that presents some advantages over traditional RL reward functions, such as being more directly informative about the optimal behaviour and easier to implement \cite{knox_training_2013}. 
Several methods have been proposed for shaping with human-provided evaluative feedback.
In the standard method, reward shaping, evaluative feedback is converted into numerical values that are used for augmenting a predefined reward function \cite{isbell_social_2001,thomaz_reinforcement_2006-1,tenorio-gonzalez_dynamic_2010,mathewson_simultaneous_2016}.
So, in reward shaping, evaluative feedback is considered in the same way as standard MDP rewards.

However, several authors have pointed out the difference between the nature of evaluative feedback and MDP rewards, considering them as information about the policy \cite{ho_teaching_2015}.
Consistent with this view, policy-shaping methods use the distribution of evaluative feedback in order to infer the teacher's policy \cite{griffith_policy_2013,loftin_strategy-aware_2014}.
This policy can be then combined with another source of information such as an MDP policy.
Overall, policy-shaping methods have been shown to perform better than reward shaping 
because they do not interfere with the reward function, hence they avoid convergence problems \cite{knox_reinforcement_2012-1,griffith_policy_2013}. 

The shaping method that we use in this paper (cf. Section \ref{sec:TM}) is closely related to the ``Convergent Actor-Critic by Humans" (COACH) algorithm \cite{macglashan_2017}. 
Both methods use evaluative feedback for updating the actor of an Actor-Critic architecture. 
However, in \cite{macglashan_2017} the update term is scaled by the gradient of the policy; whereas we do not consider a multiplying factor for evaluative feedback. 
This minor difference may have important implications on the flexibility of the teaching process.   
For instance, one can predict that multiplying by the policy gradient would dampen the effect of evaluative feedback when the policy is near a local optimum (when $\pi(s,a)$ is close to 0 or 1). 
This would make more difficult for the human teacher to rectify the policy.
This could eventually happen if evaluative feedback is combined with MDP rewards.
However, this question has not been addressed by the authors.
The main focus of our framework being on shaping with (unlabeled) instructions, we keep the comparison between these two methods for future work.

\subsubsection{Shaping with instructions}
\label{shaping-instructions}
Even though evaluative feedback provides a more direct evaluation of the behaviour than reward functions, it does not solve the exploration-exploitation dilemma since the robot still needs to try different actions before performing the optimal one \cite{sutton_reinforcement_1998}.
Moreover, previous work has shown that, in addition to evaluative feedback, human teachers want to provide guidance about future actions \cite{thomaz_reinforcement_2006}.
The role of guidance is to constrain the exploration towards a limited set of actions \cite{suay_effect_2011}. 
Instruction can be viewed is a special case of guidance, where one single action is communicated for a given situation \cite{nicolescu_natural_2003,rybski_interactive_2007,pradyot_instructing_2012}.

We generally distinguish three ways of shaping with instructions.
The first one, which is referred to as guidance, consists in simply executing the communicated action. 
For example, verbal instructions can be used for guiding the robot along the task \cite{thomaz_robot_2007,tenorio-gonzalez_dynamic_2010,cruz_interactive_2015}. 
In \cite{nicolescu_natural_2003} and \cite{rybski_interactive_2007}, a  Learning from Demonstration (LfD) system is augmented with verbal instructions, in order to make the robot perform specific actions during the demonstrations.

The second approach for shaping with instructions is to integrate the information about the action within the model of the task.
In \cite{utgoff_two_1991}, the authors present a method where a teacher interactively informs an RL agent about the next preferred state.
This information can be provided by telling the agent what action to perform.
State preferences are transformed into linear inequalities that are integrated into the learning algorithm in order to accelerate the learning process.
In \cite{clouse_teaching_1992}, instructions are integrated into an RL algorithm by positively reinforcing the proposed action.

These two approaches can be combined. 
In \cite{rosenstein_supervised_2004}, the authors present an Actor-Critic architecture that uses instructions for both decision-making and learning.
For decision-making, the robot executes a composite real-valued action that is computed as a linear combination of the actor's decision and the supervisor's instruction.
Then, the error between the instruction and the actor's decision is used as an additional parameter to the TD error for updating the actor's policy.

Another alternative is to use the provided instructions for building an instruction model besides the task model.
Both models are then combined for decision-making. 
For example, in \cite{pradyot_integrating_2012}, the RL agent arbitrates between the action proposed by its Q-learning policy and the one proposed by the instruction model, based on a confidence criterion.

\subsection{Interpreting instructions}
Classically, the meaning of instructions, \textit{i.e.} the mapping between instruction signals and the robot's actions, is determined before learning the task \cite{nicolescu_natural_2003,rybski_interactive_2007,tenorio-gonzalez_dynamic_2010,thomaz_reinforcement_2006,thomaz_robot_2007,suay_effect_2011}. 
However, as instructions can be task specific, it is difficult for non-expert users to program their meaning for new tasks. 
In addition, handcoded instructions limit the possibility for different teachers to use their own preferred signals.
One way to overcome this limitation is to let the robot interpret users' instructions. 

One method for interpreting instructions consists in providing the robot with a description of the task; then making it interact with its environment in order to interpret the instructions using either demonstrations \cite{macglashan_grounding_15} or a predefined reward function \cite{branavan_reinforcement_2009,branavan_reading_2010,vogel_learning_2010}.
In this case, instructions are provided prior to the learning process, and the robot only interacts with its environment without interacting with the human.
The main goal here is only to interpret instructions, not to use them for task learning. 

Another existing method is to teach the robot to interpret continuous streams of control signals that are provided by the human teacher \cite{mathewson_simultaneous_2016}.
In contrast to the first approach, the robot interacts only with the human and not with the environment. 
But yet, the main goal is only to interpret instructions and not to use them for task learning.

A third approach consists in guiding a task-learning process by interactively providing the robot with unlabeled instructions. 
The robot simultaneously learns to interpret instructions and uses them for task learning.
For example, in \cite{grizou_robot_2013}, the robot is provided with a set of hypotheses about possible tasks and instruction meanings.
The robot then infers the task and instruction meanings that are the most coherent with each other and with the history of observed instruction signals. 

It is important to understand the difference between these different settings.
In the first two settings where the aim is only to interpret instructions, there is no challenge about the optimality or the sparsity of the provided instructions.

First, instructions cannot be erroneous as they constitute the reference for the interpretation process.
Even though these works do not explicitly assume perfect instructions, the robustness of the interpretation methods against inconsistent instructions is not investigated.
When instructions are also used for task learning, as in our work, we have to take into account whether or not instructions are correct with respect to the target task.
However, this was not investigated in other works.
In \cite{grizou_robot_2013}, only the performance under erroneous evaluative feedback is reported.

Second, instructions cannot be sparse, since the interpretation process is defined only when instructions are available. 
For instance, the existing methods for interpreting instructions using RL \cite{branavan_reinforcement_2009,branavan_reading_2010,vogel_learning_2010,mathewson_simultaneous_2016} cannot be used with sparse instructions. 
In these methods, instructions constitute the state-space over which the RL algorithm is deployed.
This assumes the existence of a contiguous MDP state-space for computing the TD error (cf. Equation \ref{eq:dt}).
However, when instructions are interactively provided during task learning, as in our work, we have to face the challenge of sparsity with respect to task states.
So, the standard RL method used in \cite{branavan_reinforcement_2009,branavan_reading_2010,vogel_learning_2010,mathewson_simultaneous_2016} cannot be used with sparse instructions.
In this paper, we propose an alternative solution where instructions are interpreted using the TD error of the task-learning process (cf. Section \ref{sec:IM}).

\section{Model}
\label{model}
In this section, we present our framework for interactively shaping a robot behaviour with unlabeled human instructions.
In this framework, learning is primarily based on an external evaluation source, such as evaluative feedback and/or a reward function. 
So, the role of instructions is only to accelerate the learning process.

\begin{figure}
\begin{center}
\includegraphics[scale=0.4]{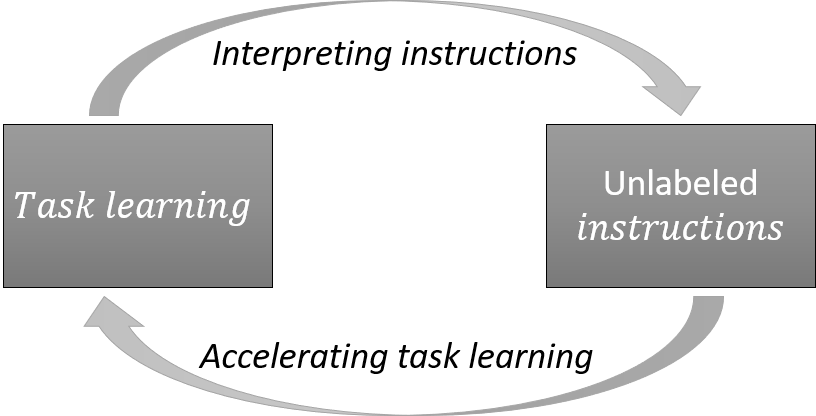}
\caption{Shaping with unlabeled instructions. Task learning is used for interpreting instructions, which are in turn used for accelerating the learning process.} 
\label{fig:instruction-task}
\end{center}
\end{figure}

Our idea is to use instruction signals as a means for sharing the information about the optimal action between different task states.
We divide this process into two sub-problems: interpreting instructions and shaping.
For interpreting instructions, the robot must find the action corresponding to each instruction signal.
For this, we take advantage of the task-learning process, to retrieve the information about the optimal action, from every task state in which each signal has been observed.
Shaping consists in using the interpreted instructions for informing the robot about the optimal action, in states where the policy has not yet converged.
These two processes are performed simultaneously and incrementally: task learning is used for interpreting instructions, and the interpreted instructions are in turn used for accelerating the learning process (Fig. \ref{fig:instruction-task}).
We first describe the general architecture of our framework. 
We then detail the methods that are implemented in each component of the architecture.
	
\subsection{The TICS architecture}
\label{model architecture}
\begin{figure}
\begin{center}
\includegraphics[scale=0.4]{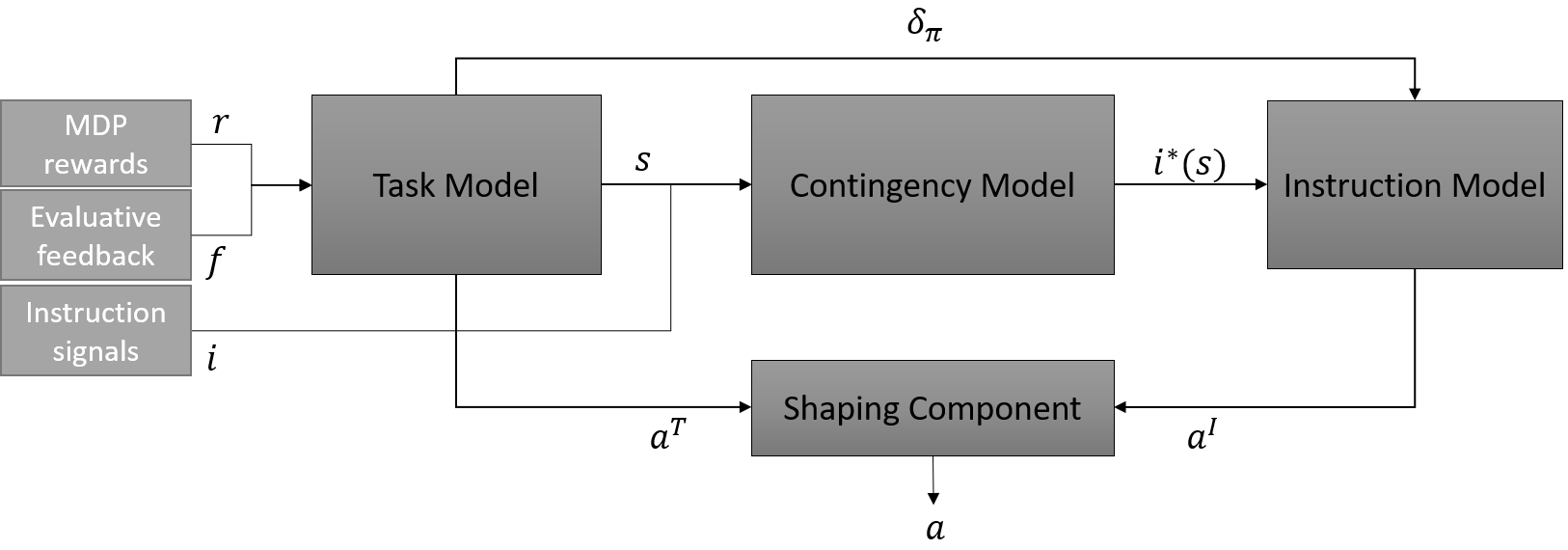}
\caption{The TICS architecture includes four main components: a Task Model learns the task, a Contingency Model associates task states with instruction signals, an Instruction Model interprets instructions, and a Shaping Component combines the outputs of the Task Model and the Instruction Model for decision-making.} 
\label{fig:archi}
\end{center}
\end{figure}

The general architecture of our framework is based on four components: a Task Model (\textsf{TM}), an Instruction Model (\textsf{IM}), a Contingency Model (\textsf{CM}) and a Shaping Component (\textsf{SC}) (Fig. \ref{fig:archi}).
We call this architecture TICS for Task-Instruction-Contingency-Shaping.
The Task Model is responsible for learning the task, while the Instruction Model is responsible for interpreting instructions. 
These two components represent the core of the TICS architecture. 
The two remaining components are meant to make the first two components interact with each other.
The Contingency Model links task states within \textsf{TM} to instruction signals within \textsf{IM}, by determining which signal has been observed in each state.
The role of this model is to minimize the number of interactions with the teacher by recalling the previously provided instructions, and also to make the mapping between states and instructions signals more robust to errors. 
Finally, the Shaping Component is responsible for combining the outputs of \textsf{TM} and \textsf{IM} for decision-making \footnote{The \textsf{TM}, \textsf{IM}, and \textsf{CM} are called Models to indicate that these are learning components, in which a model (of the task, instructions and contingency) is learned. By contrast, in the shaping component \textsf{SC}, there is no learning; the shaping method is determined in advance.}.

\paragraph{Task Model:}
The Task Model is the component of our architecture that is responsible for learning the task.
It takes as inputs task states, the performed actions and their evaluation; and it derives a task policy accordingly.
For example, when learning from a reward function, the task policy can be derived using a standard RL algorithm like Q-Learning or Actor-Critic \cite{sutton_reinforcement_1998}.
When learning from evaluative feedback, the task policy can be derived using any algorithm learning from evaluative feedback, such as TAMER \cite{knox_interactively_2009}, or the Bayesian frameworks of \cite{griffith_policy_2013} and \cite{loftin_learning_2016}.
The main challenge is to integrate different evaluation sources into one single task policy.
Combining evaluative feedback and MDP rewards is an active research question \cite{knox_reinforcement_2012-1}.
In this paper, we propose a new method for shaping with evaluative feedback, that we detail in Section \ref{sec:TM}.

\paragraph{Contingency Model:}
When the robot evaluates the task state and has to choose an action, the teacher can provide it with an instruction to indicate the optimal action to perform.
As the robot may encounter the same state several times before finding the optimal action, we want to avoid the charge for the teacher of giving the same instruction several times for the same state.

For this purpose, the Contingency Model learns a model of the co-occurrence between task states and the observed instruction signals.
This way, when the robot encounters a state for which it has already received an instruction, it can use it for interpretation and shaping.
At every step, \textsf{CM} takes as inputs the current task state and the observed instruction signal, if any. 
These inputs are used for updating the model of the contingency between states and instruction signals.
The \textsf{CM} outputs the most probable instruction signal for the current state, which is then interpreted by \textsf{IM}, and used for shaping in \textsf{SC}.

The second role of \textsf{CM} is to provide robustness against erroneous instructions by integrating the history of provided instructions, instead of considering only the signal observed in the current time-step.

\paragraph{Instruction Model:}
The Instruction Model is the unit of our architecture that is responsible for interpreting instructions.
This model takes as input from \textsf{CM} the most probable instruction signal for the current task state, in order to interpret it.
For this, it retrieves from \textsf{TM} useful information about the optimal action for the current state, and updates the meaning of instruction signals accordingly.
The challenge is to design an interpretation method that accelerates the task-learning process, while being robust against sparse and erroneous instructions. 
Our interpretation method is detailed in Section \ref{sec:IM}.

\paragraph{Shaping Component:}
The Shaping Component is responsible for integrating the information carried by \textsf{IM} into the decision-making step.
It takes as inputs information from both \textsf{TM} and \textsf{IM}, and outputs an action that would be performed by the robot.
Depending on the algorithm used for task learning and how instructions are represented, different shaping methods can be designed. 
A basic shaping method would be to consider only the action indicated by the instruction.
This would be a satisfying solution if instructions are flawless, \textit{i.e.} if the teacher always provides the correct signal and the robot always interprets it correctly.
However, in our case, we do not have such guarantees: the teacher may provide incorrect signals, and the interpretation of these signals can be incorrect. 
So, a more reasonable solution is to take into account the decisions of both \textsf{IM} and \textsf{TM}.

\subsection{Methods}
\label{methods}
The TICS architecture represents the general idea of our framework: instructions are interpreted (\textsf{IM}) according to the context in which they are provided (\textsf{TM}+\textsf{CM}) and are in turn used for guiding the learning process (\textsf{SC}).
However, each component of the architecture can be implemented in several ways.
In this section, we detail the methods we use in this paper.
First, we present how we derive a policy within the Task Model, using both evaluative feedback and MDP rewards.
Second, we explain how instructions are represented and interpreted within the Instruction Model.
Then, we describe our implementation of the Contingency Model.
Finally, we explain how the interpreted instructions are used for decision-making in the Shaping Component.

\subsubsection{Task Model}
\label{sec:TM}

For combining evaluative feedback with MDP rewards, we adopt the same view as \cite{griffith_policy_2013,ho_teaching_2015,loftin_learning_2016,macglashan_2017}, by considering feedback as information about the policy and not as standard MDP rewards.

We propose a policy-shaping method that uses evaluative feedback to incrementally update the MDP policy.
We first convert the provided feedback into binary numerical values like in standard reward-shaping methods. 
Then we use them to directly update the policy without modifying the value function.
  
\begin{figure}
\begin{center}
\includegraphics[scale=0.4]{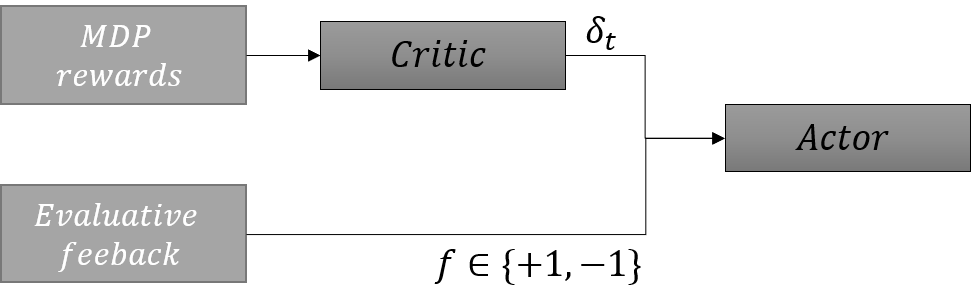}
\caption{Actor-Critic architecture. The TD error and evaluative feedback are both used for updating the actor.} 
\label{fig:actor-critic}
\end{center}
\end{figure}

For this, we use an Actor-Critic architecture, where the value function (the critic) and the policy (the actor) are stored separately (Fig. \ref{fig:actor-critic}).
When learning from MDP rewards, the actor is updated by the TD error coming from the critic using Equation (\ref{eq:ac-pu}).
When evaluative feedback $f_t\in\{-1,1\}$ is provided, it is directly used for updating the actor's parameters using:
\[p(s_t,a_t) \leftarrow p(s_t,a_t) + \beta f_{t}.\]

Like the TD error, a positive feedback increases the probability of selecting $a_t$ in $s_t$, while a negative feedback decreases it. 
This way, both feedback and MDP rewards are used for updating the same task policy, without interfering with each other.

\subsubsection{Contingency Model}
\label{sec:CM}
Storing the contingency between task states and instruction signals keeps the teacher from providing instructions several times for the same state.
However, when dealing with real robotic systems, detecting contingency between different events raises many challenges.

First, we have to make sure to detect only relevant events.
In our case, we want to ensure that the detected signals from the teacher correspond to instructions and not to any other signal.
For this, we define a set of possible instruction signals among which the teacher can choose.
Whenever one of these signals is detected, we know that the teacher is providing instruction.

But still, a detected instruction signal can be due to an error either from the teacher or from the detection system.
To limit these errors, we define contingency for a given state as a probability distribution over detected signals and we only consider the most likely one as the true instruction signal.

In this paper, we propose a simple implementation of the Contingency Model.
We consider a co-occurrence matrix that stores the number of times each signal has been detected in every state.
Whenever an instruction signal $i$ is detected in a state $s_t$, the number of co-occurrences $c(s_t,i)$ of $s_t$ and $i$ is incremented.
Then, we update the Contingency Model of $s_t$ according to the newly detected signals using

\[Pr(i|s_t) = \frac{c(s_t,i)}{\sum_{j \in I} c(s_t,j)}.\]

The most likely instruction signal $i^*$ for $s_t$ is the one that has been observed most of the time in this state:
\[i^*(s_t) = \argmax_{i \in I } Pr(i|s_t).\]

\subsubsection{Instruction Model}
\label{sec:IM}
Interpreting instructions consists in finding the optimal action corresponding to each instruction signal.
To do so, we retrieve information about the optimal action from the task-learning process.
We first explain how we represent instructions, then we detail our interpretation method.

\paragraph{Representing instructions}

In this paper, we represent instructions as a probability distribution over actions.
One interesting feature of this representation is that it enables incremental updates and it naturally represents different hypotheses about the meaning of an instruction signal.
In addition, it is mathematically equivalent to the definition of a state policy, which is also interesting in terms of (policy) shaping.
In the remainder, we refer to the model of an instruction as a \textit{signal policy}, in contrast to the notion of \textit{state policy}. 
We denote the signal policy for an instruction signal $i \in I$ at time $t$ as
\[\pi_t(i) = \{\pi_t(i,a); a \in A\} = \{Pr_t(a|i); a \in A\}.\]

\paragraph{Interpreting instructions}

The most straightforward method for computing a signal policy $\pi_t(i)=\{ \pi_t(i,a) ; \forall a \in A\}$ is to consider instruction signals as the state-space of an alternative MDP, which can be solved using a standard RL algorithm.
Since each instruction signal is sufficient for indicating the optimal action, state transitions in this MDP satisfies the Markov property.
However, this is only true if the teacher provides an instruction signal for every state.
In fact, RL algorithms require contiguous state transitions for computing the TD error (cf. Equation \ref{eq:dt}).
Since we consider an interactive learning scenario, this condition is not necessarily true.
The teacher may not provide instructions for every situation.
So, the main limitation of this interpretation method is that it is valid only with non sparse instructions.
This approach has been previously considered in the literature \cite{branavan_reinforcement_2009,branavan_reading_2010,vogel_learning_2010,mathewson_simultaneous_2016}. 
We refer to this method as \textbf{Reward-based updating (RU)}.

In this paper, we propose an alternative solution where instructions are interpreted using the TD error of the Task Model.
We update a signal policy with the same amount as its corresponding state policy, the idea being that an instruction signal indicates the optimal action of its corresponding task state.
When the robot visits a state $s_t$ such as $i^*(s_t)=i$ and performs action $a_t$, it updates the signal policy $\pi_t(i,a_t)$ using:
\[\delta \pi_t(i,a_t)=\delta \pi_t(s_t,a_t).\]

In practice, the update is performed through the parameters of both policies using:
\[\delta p_t(i,a_t)=\delta p_t(s_t,a_t).\]
We refer to this interpretation method as \textbf{Policy-based updating (PU)}.
In contrast to RU, PU can be used with sparse instructions since it does not require contiguous instruction signals to compute the TD error.

\subsubsection{Shaping Component}
\label{model-shaping}

The Shaping Component combines the policies of the Task Model and the Instruction Model, in order to output an action to be executed by the robot. 
In Section \ref{shaping-instructions}, we presented some existing methods for shaping with instructions.

In this paper, we use the method proposed by \cite{pradyot_integrating_2012}, where a confidence measure is used for arbitrating between several policies.
Let $x$ denote either a task state $s \in S$ or an instruction signal $i \in I$. 
The confidence of a policy $\pi$ on $x$ is given by
\[ \kappa_{\pi}(x) = \max_{a \in A} \pi(x,a) - \max_{b \neq a} \pi(x,b).\]

This measure reflects the certainty of a policy and takes values between $0$ and $1$.
For instructions, it reflects the confidence about the interpretation of an instruction signal.
At time-step $t$, if $\kappa_{\pi}(s_t) < \kappa_{\pi}(i^*(s_t))$, the decision is taken according to the Instruction Model.
Otherwise, it is taken according to the Task Model.

\section{Simulation protocol}
\label{protocol}
In this section, we present the experimental protocol that we use for evaluating our framework in simulation.
We first introduce the problem domains that we consider.
Second, we describe the teaching protocol used for providing teaching signals in the simulated environments.
Then, we detail our evaluation criteria.

\subsection{Problem domains}
\label{domains}
We evaluate our framework on two different problems: object sorting and maze navigation.
Our motivation is to evaluate our framework on different MDP structures, in order to make our results more general.
The contrasting properties of these two domains have different implications on the performance of our model.
So, we introduce these two problems and we highlight the difference between their structures.

\subsubsection{Object sorting}
\begin{figure}
\centering
\subfloat[Object sorting domain. A robot facing a table divided into three regions: $z_1$, $z_2$ and $z_3$. A camera (red rectangle) is used for extracting the visual features of an object placed in $z_2$. We consider two types of objects: $Plain$ (left) and $Pattern$ (right), with two different sizes and three colors. A Kinect sensor (blue rectangle) extracts feedback and instruction signals from the teacher.]{\includegraphics[width=0.48\linewidth]{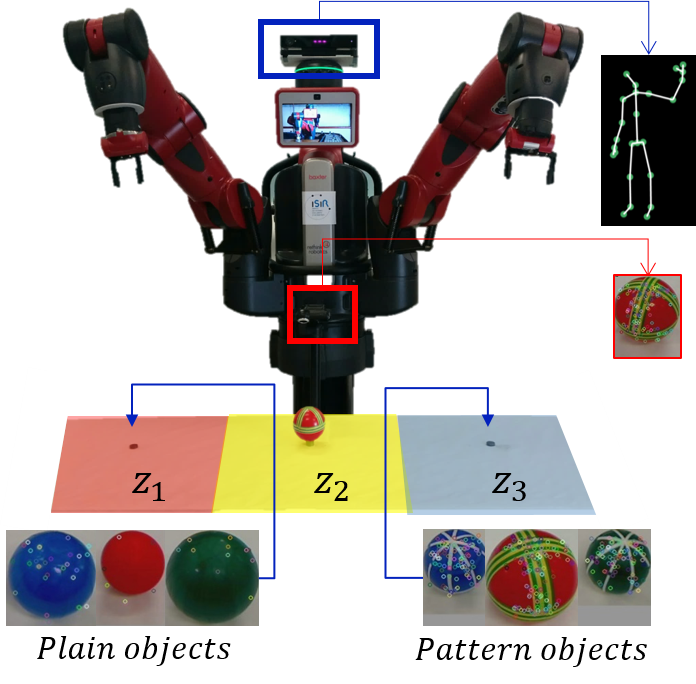}\label{fig:scenario-a}}
\hfill     
\subfloat[Maze navigation domain. Standard (top) and simplified (bottom) maps. White squares represent the free positions in the maze. The yellow square represents the goal state. The arrows represent the teacher's policy.]{\includegraphics[width=0.48\linewidth]{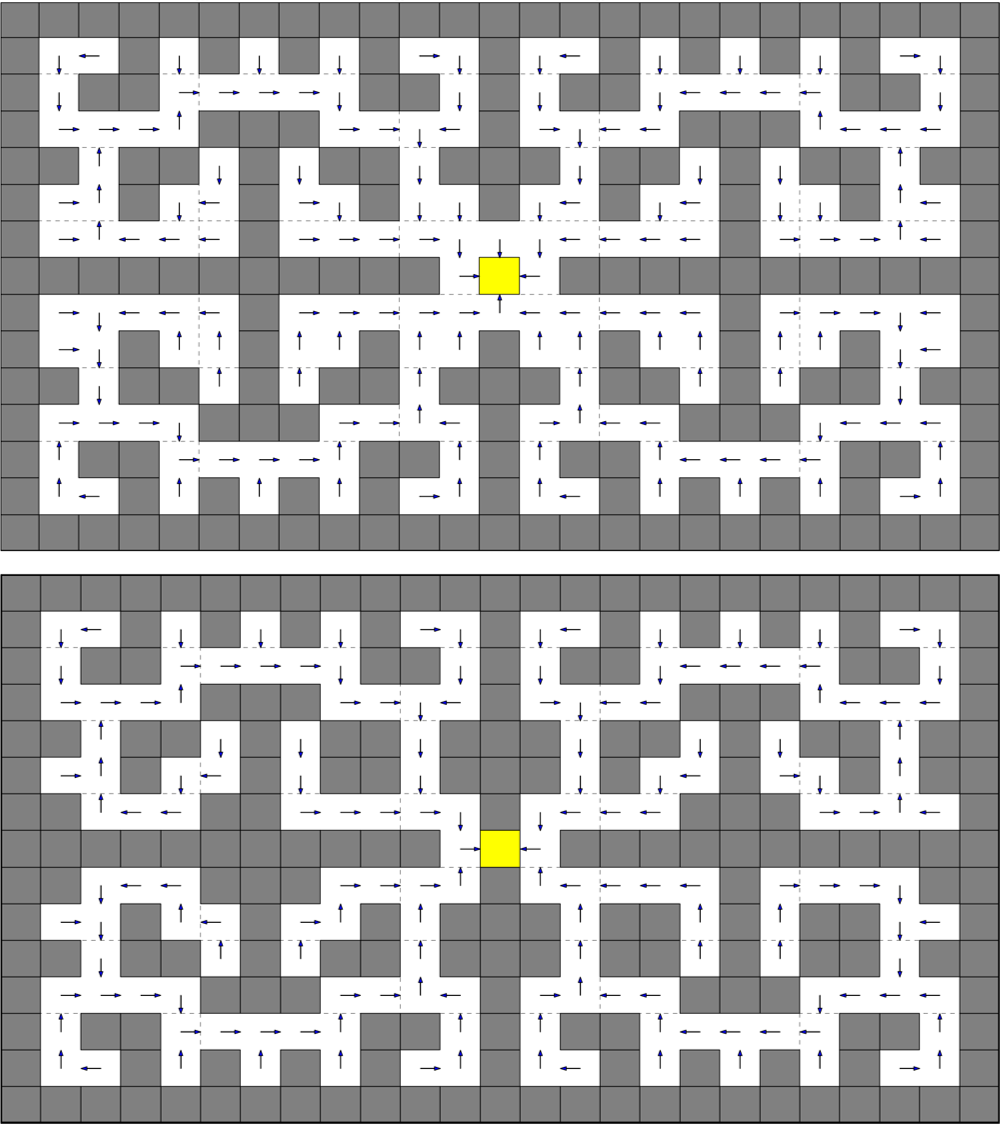}\label{fig:scenario-b}}
\caption{The problem domains used in simulation: object sorting and maze navigation. These domains have contrasting MDP structures. a) Object sorting has a clustered  state-space and a relatively large action-set (9) compared to its goal-horizon (4). b) Maze navigation, has a fully connected state-space and relatively small action-set (4) compared to its goal horizon (24).}
\label{fig:scenario} 
\end{figure}

We consider the object sorting domain introduced in \cite{suay_effect_2011}\footnote{ 
We also consider this task for the experiment with the real robot (cf. Section \ref{baxter}).}. 
In this task, the robot has to sort different objects according to their visual features. 
The scenario (Fig. \ref{fig:scenario-a}) involves a robot facing a table on top of which we can place objects in three different positions: $z_1$, $z_2$ and $z_3$.
When an object is presented to the robot in $z_2$, it must pick it up and then place it in the appropriate zone according to its type. 
Unicolor objects ($Plain$) must be placed in $z_1$ while objects with patterns ($Pattern$) must be placed in $z_3$.
We consider eight different objects in our simulated task.

The task is defined as an MDP $<S,A,T,R>$.
The state space $S$ is defined by the tuple $S = (L_{lh}, L_{rh}, L_o, D)$. 
$L_{lh} = \{z_2, z_3\}$ and $L_{rh} = \{z_1, z_2\}$ represent the locations of the robot's hands.
Each hand can be located in two different positions above $z_1$, $z_2$ and $z_3$. $L_o = \{z_1, z_2, z_3, lh, rh\}$ describes the position of the object which can be located in one of the three zones or in one of the robot's hands. 
Finally, $D= (TYPE, COLOR, SIZE)$ describes the visual features of the object: $TYPE \in \{Plain, Pattern\}$ describes its type, $COLOR \in \{Red, Green, Blue\}$ describes its dominant color, and $ SIZE \in \{Large, Small\}$ describes its size. 

The robot is able to perform nine elementary actions that are necessary for completing the task $A = \{TakePicture, xMoveLeft, xMoveRight, xPick,$ $ xPlace\}$, where $x \in \{LeftHand,$ $ RightHand\}$. The $TakePicture$  action takes an image of $z_2$ and extracts its visual features. 
We do not consider this action in simulation, which means that the descriptors of the object are always known to the robot, and the action space contains only eight actions.

To make the robot learn the task, we divide each training session into several episodes.
Each episode begins with the teacher placing an object in $z_2$, and terminates whenever the robot places the object either in the correct or the wrong spot, which leads to either a success or a failure.
When learning from a predefined reward function, the robot gets a reward $r=-0.1$ for every transition until the end of the episode.
A successful episode gives to the robot a positive reward $r=1$, while a failure gives a negative reward $r=-1$.
We consider a deterministic transition function.

The object sorting domain has some particular characteristics with respect to standard RL problems.
First, the state-space is not connected, as the reachable states in each episode are determined by the object placed on the table.
Second, the size of the action-set is relatively high (9) compared to the goal horizon (4).

\subsubsection{Maze navigation}
Maze navigation is a standard problem in the RL literature, where a robot must navigate within a maze in order to reach a predefined goal state (Fig. \ref{fig:scenario-b}). 
The task is represented as an MDP  $<S,A,T,R>$.
The state space $S$ is defined by the coordinates of each state $S = (s_x,s_y)$.
The robot can perform four elementary actions $A = \{North, East, South,$ $ West\}$ that enable it to navigate through the maze.
Each training session is divided into several episodes.
At the beginning of each episode, the robot's location is randomly initialized in a free position.
The episode ends whenever the robot reaches the goal state.

In our experiments, we consider two different instances of this problem.
In the standard maze problem, some states can have multiple optimal actions; while in the simplified maze problem, there is only one optimal action in every state.
The existence of multiple optimal actions has implications on the interpretation of instructions.
When learning from a predefined reward function, the RL algorithm can converge to a policy that is different from the one being communicated by the teacher.
As interpretation is based on the task-learning process, some instructions can be misinterpreted because of this ambiguity.

We consider a deterministic transition function for this problem.
The reward function is defined by a negative reward $r=-0.01$ for every intermediate transition and a positive reward $r=1$ for reaching the goal.

Unlike object sorting, the maze navigation domain has a fully connected state-space as all states are reachable in every episode. In addition, the size of its action-set is relatively small (4) compared to the goal horizon (24).
Most importantly, in maze navigation, some states can have several possible optimal actions, as noted above.

\subsubsection{Parameters}
Table \ref{tab:params} reports the parameters used in our experiments.
We use the same parameters for all models in both domains.
To evaluate the performance of a model on a given domain, we run $1000$ training sessions.
Each session involves $1000$ consecutive episodes with $1000$ maximum steps each.
The horizon parameter $\gamma$ is set to $0.9$ and both learning rates $\alpha$ and $\beta$ are set to $0.1$.
We use an $\epsilon-$greedy action-selection strategy with $\epsilon=0.1$, and a decay parameter for $\epsilon$ after each step of $\delta_\epsilon = 0.001$ ($\epsilon$ reaches $0$ after $100$ steps). 
The value function is initialized to $0$, at the beginning of every training session.

\begin{table}
\caption{Experiment parameters.} 
\centering
\begin{tabular}{|c|c|c|c|c|c|c|c|}
\hline
$|$exp$|$ & $|$episodes$|$& max steps & $\gamma$	& $\alpha$ & $\beta$ & $\epsilon$ &	$\delta_{\epsilon}$ \\
\hline
$1000$ & $1000$ & $1000$ &	$0.9$ 	& $0.1$	       & 	$0.1$	&	    $0.1$ 	&	$0.001$ \\
\hline
\end{tabular}
\label{tab:params} 
\end{table}

\subsection{Simulated teacher}
\label{simu-protocol}
For each task, we assume that the simulated teacher has one single preferred optimal policy $\pi_T^*$ that it tries to communicate to the robot.
Evaluative feedback and instructions are provided according to this policy.
For instructions, the teacher uses one single instruction signal $i(a)$ to indicate each action $a \in A$.
For every state $s \in S$, it provides the instruction signal $i(\pi_T^*(s))$ corresponding to its own optimal policy.
If the robot executes the action $a_t = \pi_T^*(s_t)$, the teacher provides it with positive feedback $fb_t=1$. 
Otherwise, it provides a negative feedback $fb_t=-1$.
We note $fb_T^*(s_t,a_t)$ the optimal feedback according to the teacher's preferred policy: 

\begin{equation}
    fb_T^*(s_t,a_t)=
    \begin{cases}
      1, & \text{if}\ a_t = \pi_T^*(s_t) \\
      -1, & \text{otherwise.}
    \end{cases}
\end{equation}

We assume a transparency-based protocol for providing instructions, where the teacher has access to the output of the Contingency Model in every step \cite{thomaz_transparency_2006}.
So, the teacher knows whether or not an instruction signal has been associated to the current state and whether or not the associated signal is correct, and provides instruction only if no signal is associated to the current state, or if the displayed signal is incorrect.
The main motivation behind this protocol is to limit the number of interactions with the robot, so the teacher provides instruction only when required.
For feedback, we assume the teacher provides evaluative feedback for every performed action.

We evaluate our model under three teaching conditions: ideal, sparse and erroneous teaching signals. 
In the ideal case, the teacher provides instructions for every state $s \in S$ and feedback for every state-action pair $(s,a) \in S \times A$.
In the sparse condition, instructions are provided only in a subset of the state-space ($Pr(i|s) < 1$) and feedback is provided only for a subset of state-action pairs ($Pr(fb|s,a) < 1$).

In the erroneous case, when the teacher provides instructions, there is a probability that the robot perceives a random instruction signal $i_t$ that is different from the one corresponding to the teacher's preferred action: $i_t \neq i(\pi_T^*(s_t))$. For erroneous feedback, we consider that at every step, the robot may perceive an evaluative feedback that is inconsistent with the teacher's policy: $fb_t \neq fb_T^*(s_t,a_t)$.

\subsection{Evaluation criteria}
To compare different models, we use two evaluation criteria: the convergence rate of task performance and the interaction load of the teaching process.

\subsubsection{Convergence rate}
As instructions are employed to speed up the learning process, we compare different models using the convergence rate of their task performance.
To do so, we define a convergence criterion for each problem, and we report the number of learning steps that are needed to reach it.

For the object sorting task, we assume that a session converges whenever the robot is able to perform the task in exactly $4$ steps per episode.
For maze navigation, as initial states have different distances from the goal, we consider the highest distance from the goal as a convergence criterion. 
So, we assume that a session converges whenever the robot is able to reach the goal in less than $24$ steps.

The convergence rate of each model is computed as follows.
First, we tally the number of steps over the episodes until convergence, for each training session. 
Then, we compute the density histogram of this measure over all sessions, and convert it into a cumulative distribution function.
This provides the probability that task performance converges before $n$ steps.
If this probability reaches the value $1$ for $n$ steps, it means that the model performance converges within at most $n$ learning steps.

While this measure may provide only a rough idea about convergence over a training session (\textit{e.g.} in maze navigation), it is sufficient for comparing the convergence rates of different models, as long as we use the same criterion.
The main advantage of this method is to provide a compact measure of the convergence rates over all sessions.
For a matter of compactness, in this paper we only report the 99th percentile of the number of steps until convergence, which constitutes a worst-case evaluation criterion, and the number of sessions over 1000 that did not converge.
Worst-case evaluation provides a better estimation of the reliability of a system compared to average performance (cf. figures 1 and 5 in supplementary material). 
Some cumulative distributions are also provided as supplementary material. 

\subsubsection{Interaction load}
\label{load-definition}
The second evaluation criterion is the interaction load of the teaching process, that we measure as the number of required teaching signals (feedback and instructions) for learning the task.
We report the maximum number of required teaching signals over 1000 training sessions as a worst-case evaluation criterion.

When learning from evaluative feedback only, without using instructions, the total number of required teaching signals equals the number of provided feedback until convergence of task performance. 
When instructions are used in addition to feedback, however, the total number of teaching signals required for learning is not related to the convergence of task performance.
Indeed, task performance mainly depends on the complexity of the state-space that is defined by instruction signals; and converges whenever the robot learns to interpret instructions.
From that time on, the robot only needs to use the Instructions Model, in states where instructions have been provided.
But there may exist some states that did not receive any instruction, and where the policy did not yet converge.
In this case, the teacher still needs to provide instructions for these states.

So, when using instructions, the number of required teaching signals in a session equals the number of evaluative feedback until the convergence of task performance, plus the number of instructions provided over the whole session.
Here, the number of evaluative feedback is the one required for interpreting instructions; and the number of instructions is the one required for shaping the policy, or more exactly for entirely determining the Contingency Model.

\section{Results in simulation}
\label{results}
In this section, we evaluate our framework in simulation under different hypotheses about the teaching conditions. 
First, in Section \ref{ideal}, we evaluate our model under ideal teaching signals.
Sections \ref{gsparse} and \ref{gerr} report the performance of our model with respectively sparse and erroneous instructions.
Sections \ref{fbsparse} and \ref{fberr} evaluate the robustness of our model against sparse and erroneous feedback.
Finally, in Section \ref{interaction-load}, we evaluate the cost of our method in terms of interaction load.

\subsection{Ideal teaching signals}
\label{ideal}

\begin{figure}
\centering
\subfloat[Object sorting.]{\includegraphics[trim=0.7cm 0.5cm 1.5cm 1cm,  clip=true,width=0.49\linewidth]{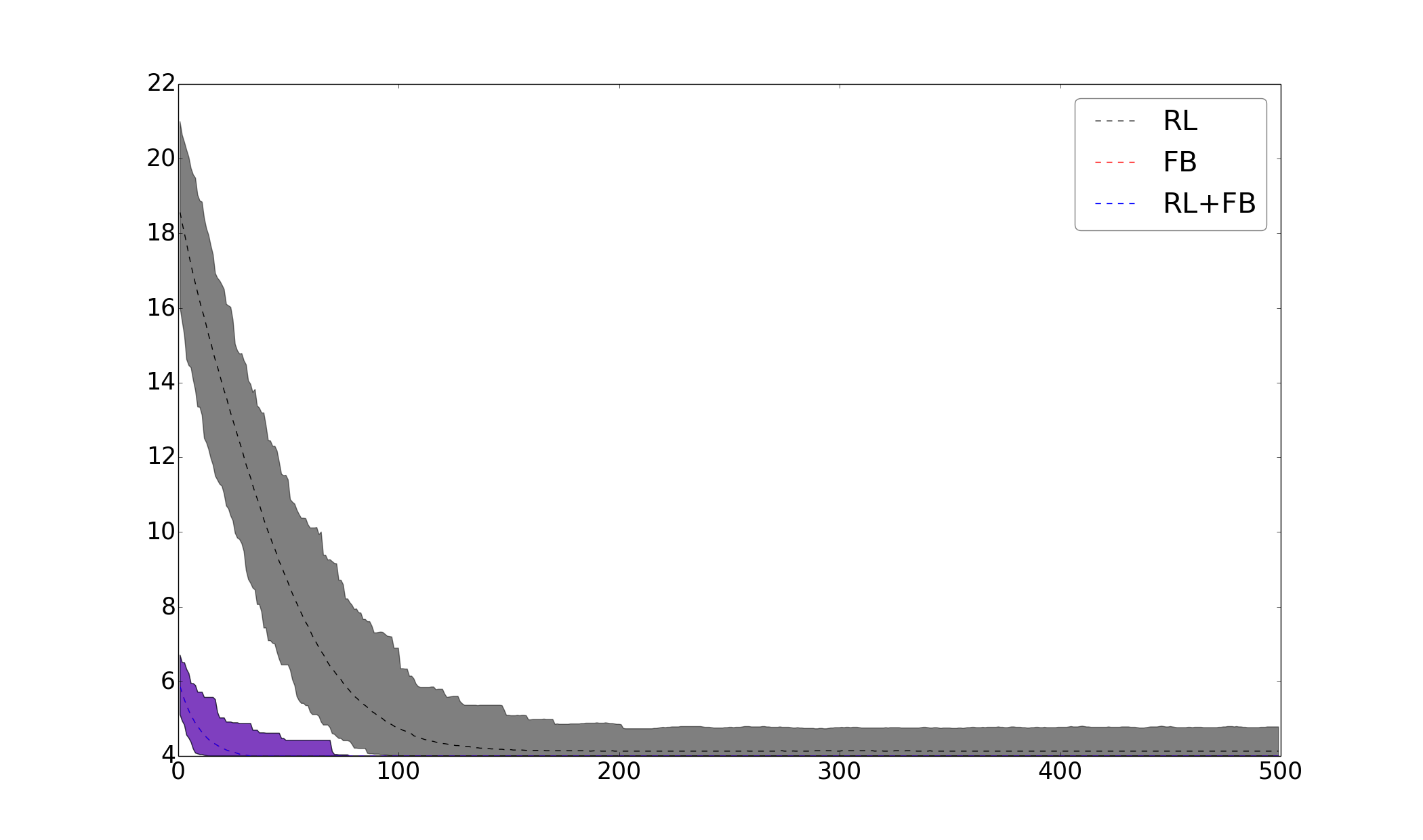}\label{fig:ideal-perfs-a}}
\hfill     
\subfloat[Maze navigation.]{\includegraphics[trim=0.7cm 0.5cm 1.5cm 1cm,  clip=true,width=0.49\linewidth]{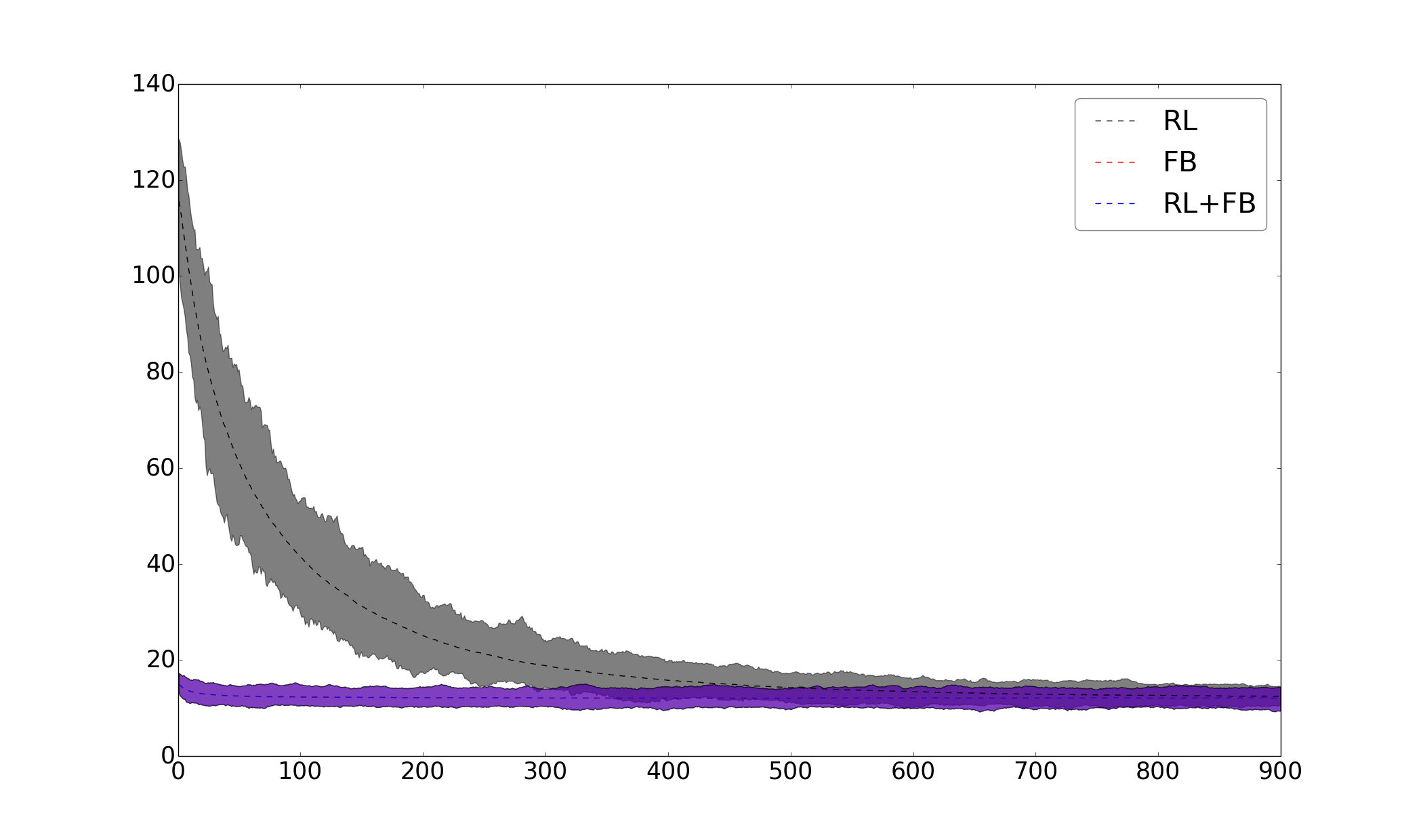}\label{fig:ideal-perfs-b}}
\caption{Performance of the baseline models in the ideal case. Each curve represents the evolution of the number of steps by episode over a training session.
For each episode, we report the median, the minimum and the maximum number of steps over 1000 sessions. 
Each curve is smoothed using a moving average over a window of $100$ episodes. 
In black: learning from a predefined reward function. In red: learning from evaluative feedback. In blue: learning from both evaluation sources. Blue and red curves completely coincide (purple), which means that the effect of the reward function is negligible compared to evaluative feedback.}
\label{fig:ideal-perfs} 
\end{figure}

Under this condition, we assume that the simulated teacher is always available and always provides correct instructions and feedback, according to its preferred policy.

We first report the performance of the three baseline models in both problem domains (Fig. \ref{fig:ideal-perfs}).
We refer to the baseline model that learns only from evaluative feedback as FB, the one that learns only from a reward function as RL, and the model that learns from both evaluation sources as RL+FB. 
None of these models use instructions.
However, they use the same Actor-Critic implementation as the Task Model.

First, we observe that in the ideal case, FB converges much faster than RL.
This is mainly due to the difference between evaluative feedback and the reward function in terms of sparsity and temporal credit-assignment. 
While evaluative feedback informs about every performed action, the effect of the reward function needs to be iteratively propagated over the entire state-action space.
This also explains the second observation, that RL+FB is equivalent to FB. 
Evaluative feedback dominates over the reward function when they are combined.

Second, we note that for the object sorting domain, the RL algorithm does not always converge to the solution of $4$ steps by episode (Fig \ref{fig:ideal-perfs-a}). 
However, the robot still converges to a solution within at most $5$ steps.
At worst, it performs one additional and useless action, like changing the position of the empty hand before putting the object in the correct zone.
Consequently, when necessary, we also consider the 5-steps by episode convergence criterion in addition to the 4-steps criterion for measuring the convergence rate on the object sorting task.

\begin{table}
\caption{Comparison of interpretation methods with ideal teaching signals. The reported numbers correspond to the 99th percentile of the number of steps until convergence. In parentheses, the number of sessions over 1000 that did not converge. }
\small
\tabcolsep=0.12cm
\begin{tabular}{|l|ccc|}
\hline

&\multicolumn{3}{c|}{Learning from evaluative feedback}\\
   & PU & RU & FB\\ 
sorting  & \textbf{102}  &  \textbf{102}  & 503\\
maze     & \textbf{38}   & \textbf{38}    & 393\\
\hline
&\multicolumn{3}{c|}{Learning from a predefined reward function}\\
   & PU & RU & RL\\
sorting & \textbf{1248}  & 2881 & 2476 (716)  \\
maze    &  \textbf{9141} & 5.2e+4  & 2.5e+4  \\
\hline 
\end{tabular}
\label{tab:ideal-1} 
\end{table}

We now assess the performance of our interpretation method, Policy-based Updating (PU), with respect to Reward-based Updating (RU) and to the baselines (Table \ref{tab:ideal-1}, Fig. 1 in supplementary material).
For this, we only consider the output of the Instruction Model in decision-making, so the performance of each interpretation method only depends on the quality of its interpretation.
When learning from evaluative feedback, we observe that both interpretation methods have the same performance; and both of them improve the convergence rate compared to FB, in both domains (for RU, evaluative feedback is converted into rewards).
When learning from a predefined reward function, however, PU outperforms RU in both domains.
Even though RU correctly interprets instructions (the Instruction Model converges to the optimal policy), it does not improve the convergence rate compared to RL.

\begin{table}
\caption{Model performance under ideal teaching signals. The reported numbers correspond to the 99th percentile of the number of steps until convergence. In parentheses, the number of sessions over 1000 that did not converge. }
\small
\tabcolsep=0.32cm
\begin{tabular}{|l|cccccc|}
\hline
&\multicolumn{6}{c|}{Learning from evaluative feedback and/or a predefined reward function}\\
  & FB+PU & RL+FB+PU & FB & RL+FB & RL+PU & RL \\ 
sorting  & \textbf{103}  & \textbf{103}  & 503  & 503  & 1298  & 2476 (716)\\
maze  & \textbf{39}  & \textbf{39}  & 393  & 393 & 3011  & 2.5e+4\\
\hline 
\end{tabular}

\label{tab:ideal-2} 
\end{table}

In Table \ref{tab:ideal-2}, we assess the performance of our full model using the PU method with respect to the baselines.
We observe that in both domains the best performance is obtained with FB+PU, namely when unlabeled instructions are used in addition to evaluative feedback. 
The second best performance is achieved when only using evaluative feedback (FB). 
Then, we have the performance obtained when unlabeled instructions are used in addition to a predefined reward function (RL+PU).
Finally, the worst performance is obtained when only using a reward function.
We can see that using unlabeled instructions with evaluative feedback accelerates the learning process by $80\%$ in object sorting and $90\%$ in maze navigation.
When learning from a predefined reward function, the learning process is accelerated by $48\%$ in object sorting and $88\%$ in maze navigation.

From these results, we conclude that with ideal teaching signals, our framework improves the convergence rate with respect to not using unlabeled instructions.

\subsection{Sparse instructions}
\label{gsparse}
We now evaluate our framework with sparse instructions.
Under this condition, instructions are provided only in a subset of the state space, while feedback is provided at every step.

\begin{table}
\caption{Model performance under sparse instructions. The probability of receiving an instruction over the entire state space is $p=Pr(i|s)$. The reported numbers correspond to the 99th percentile of the number of steps until convergence. In parentheses, the number of sessions over 1000 that did not converge. (*) Convergence to 5 steps by episode criterion. (**) Simplified maze problem.}
\small
\tabcolsep=0.1cm
\begin{tabular}{|l|ccccccc|c|}
\hline
  & p=1 & p=0.9 & p=0.7 & p=0.5 & p=0.3 & p=0.1 & p=0 & baseline\\ 
  \hline
&\multicolumn{7}{c|}{FB+PU} & FB\\
sorting  & 103  & 278  & 363  & 430  & 483  & 516  & 503 & 503 \\
maze  & 39  & 48  & 60  & 113  & 207  & 349  & 393 & 393 \\
\hline 
&\multicolumn{7}{c|}{RL+PU}& RL\\
sorting  & 1298  & 1452 (234) & 1738 (506) & 1995 (630) & 2334 (702) & 2354 (741) & 2476 (716) & 2476 (716)\\
sorting5*  & 989  & 1136  & 1428  & 1751  & 1981  & 2128  & 2157 & 2157 \\
maze  & 3011  & 1.4e+4  & 3e+4  & 6.7e+4  & 9.9e+4 (74) & 5.7e+4 (219) & 2.5e+4 & 2.5e+4 \\
maze2**  & 3555 & 9692  & 1.5e+4  & 1.9e+4  & 2.2e+4  & 2.3e+4  & 2.4e+4 & 2.4e+4 \\
\hline 
&\multicolumn{7}{c|}{RL+FB+PU}& RL+FB\\
sorting  & 103  & 278 & 363  & 430  & 483  & 516  & 503 & 503 \\
maze  & 39  & 48  & 60 & 113  & 207  & 349  & 393 & 393 \\
\hline 
\end{tabular}

\label{tab:sparse-inst} 
\end{table}

When learning from evaluative feedback, we observe the same behaviour for both problem domains (Table \ref{tab:sparse-inst}, Fig. 2 in supplementary material).
With no instructions, our model is equivalent to the baseline.
The more the teacher provides instructions for different states, the higher the probability for our model to converge faster.
These results demonstrate the robustness of our method against sparse instructions, in the case of learning from evaluative feedback.

When learning from a predefined reward function, however, this behaviour holds only for the object sorting task (Table \ref{tab:sparse-inst}, Fig. 3 in supplementary. material). 
In maze navigation, our model is not robust beyond a certain level of sparsity $Pr(i|s) \leq 0.5$. 
This issue can be explained by the existence of multiple optimal policies, since many states can have two optimal actions.
For high instruction sparsity, there is more risk that a signal is given in states where the optimal action found by the Task Model (RL) is different from the teacher's policy. 
In this situation, the meaning of the provided instruction signal can be misinterpreted by the robot, as it might be associated with another action than the one intended by the teacher. 

We verify this on the simplified maze navigation domain which has a unique optimal policy (maze2 in Table \ref{tab:sparse-inst}, also see Fig. 3 in suppl. material). 
In this domain, our model is robust against all levels of instruction sparsity and its performance increases proportionally to the number of provided instruction signals.

For domains with multiple optimal policies, different solutions can be considered to avoid the misinterpretation of instructions.
In our work, we assumed for simplicity that the teacher has one single preferred policy, according to which (s)he provides instructions.
This corresponds to the special case where the teacher knows only one way of performing the task.
In a real-world scenario, however, if the teacher is aware about the different optimal actions, (s)he may alternate between different instruction signals.
This would lower the effect of misinterpretation as the provided instruction signals would match more often with the optimal action found by the Task Model; so the robot would not get stuck into one wrong interpretation. 
In the extreme case, the teacher could change its policy to match the robot's actions.
Otherwise, if the teacher wants to persist on its own policy, (s)he can always correct misinterpretations through evaluative feedback, forcing the robot to follow the communicated policy.

This is the solution that we consider in our work.
Table \ref{tab:sparse-inst} shows that when the reward function is combined with evaluative feedback, our model becomes totally robust to sparse instructions.
As the effect of evaluative feedback dominates over the effect of the reward function, correcting misinterpreted instructions can be done easily.
This is particularly true with the method we use for interpretation, PU, which is based on the TD error of the Task Model.
Because this error tends to zero as the robot learns, the effect of the reward function on interpretation fades over time.
Consequently, there is a certain point in time where the effect of corrections cannot be overridden by the reward function.

In summary, these results show that our framework is robust against sparse instructions, when learning from evaluative feedback.
Learning from a reward function, however, may cause convergence issues in domains with multiple optimal policies, due to misinterpreting instructions.
Finally, when learning from both evaluation sources, evaluative feedback provides robustness against sparse instructions by allowing the teacher to correct misinterpreted instructions.

\subsection{Erroneous instructions}
\label{gerr}

We now evaluate the robustness of our model against erroneous instructions.
Under this condition, instructions and feedback are not sparse.
However, each time the teacher provides an instruction, we include a probability for the robot to perceive a signal that is inconsistent with the teacher's policy.

\begin{table}
\caption{Model performance under erroneous instructions. The probability of receiving an erroneous instruction is $p = Pr(i_t \neq i(\pi_T^*(s))$. The reported numbers correspond to the 99th percentile of the number of steps until convergence. In parentheses, the number of sessions over 1000 that did not converge. (*) Although the 99th percentile is higher than the baseline, the model still outperforms the baseline in more than $90\%$ of the sessions (See Fig. 4 in supplementary material). }
\small
\tabcolsep=0.1cm
\begin{tabular}{|l|ccccccc|c|}
\hline
   & p=0 & p=0.1 & p=0.3 & p=0.5 & p=0.7 & p=0.9 & p=1 & baseline \\
   \hline
&\multicolumn{7}{c|}{FB+PU}&FB\\
sorting    & 103  & 329  & \textbf{568*}  & 1011  & 3056 (2) & 4096 (294) & 3974 (65) & 503\\
maze   & 39  & 49  & \textbf{253}  & 1756  & 1.8e+4  & 1.4e+4  & 1.4e+4 & 393  \\
\hline 
&\multicolumn{7}{c|}{RL+PU}&RL\\
sorting   & 1298  & 1316  & 1543  & \textbf{2014}  & 3907 (3) & 5034 (995) & 6240 (987)& 2476 (716) \\
maze    & 3011  & 5325  & 11566  & \textbf{2.4e+4}  & 1.6e+5 (4) & NA & NA & 2.5e+4\\
\hline 
&\multicolumn{7}{c|}{RL+FB+PU}&RL+FB\\
sorting    & 103  & 340  & \textbf{512*}  & 1065  & 2904 (1) & 4201 (258) & 3964 (63) & 503\\
maze    & 39  & 52  & \textbf{259}  & 1784  & 1.9e+4 (1) & 1.4e+4  & 1.4e+4 & 393 \\
\hline 
\end{tabular}

\label{tab:err-inst} 
\end{table}

Generally, we observe that our model is robust to erroneous instructions up to a probability of error $Pr(i_t \neq i(\pi_T^*(s_t)) \leq 0.3$  (Table \ref{tab:err-inst}).
When learning from evaluative feedback, our model still outperforms the baseline for a probability of error $Pr(i_t \neq i(\pi_T^*(s_t)) \leq 0.3$, in both domains. 
In the object sorting domain, our model outperforms the baseline in $97\%$ of sessions (see Fig. 4 in supplementary material).
When learning from a predefined reward function, our model outperforms the baseline for $Pr(i_t \neq i(\pi_T^*(s_t)) \leq 0.5$ in both domains. 
Combining the reward function with evaluative feedback does not provide more robustness against erroneous instructions and the combination has almost the same performance as when learning only from evaluative feedback.

These results show that our framework is robust against erroneous instructions and improves the convergence rate, for a probability of erroneous instructions lower than $0.3$.

\subsection{Sparse feedback}
\label{fbsparse}

One limitation of learning from evaluative feedback is that the quality of the learning process mainly depends on the number of provided feedback.
If feedback is too sparse, the learning process can be penalized.
So, we propose to evaluate the performance of our framework under different levels of feedback sparsity.
We consider the condition where instructions are not sparse, while feedback is provided only for a subset of state-action pairs.

We first consider the situation where evaluative feedback is the only available evaluation source (Table \ref{tab:fb-sparse}). 
In this case, we observe that the FB+PU model makes the learning process more robust to higher levels of feedback sparsity.
In the object sorting task, the performance of the baseline model FB is drastically decreased whenever feedback becomes sparse ($Pr(fb|s,a) < 1$). 
With FB+PU, however, the learning process is still robust down to a probability of feedback $Pr(fb|s,a) \geq 0.7$.

\begin{table}
\caption{Model performance under sparse feedback. The probability of receiving a feedback over the entire state-action space is $p=Pr(fb|s,a)$. The reported numbers correspond to the 99th percentile of the number of steps until convergence. In parentheses, the number of sessions over 1000 that did not converge.}
\small
\tabcolsep=0.1cm
\begin{tabular}{|l|ccccccc|c|}
\hline
& p=1 & p=0.9 & p=0.7 & p=0.5 & p=0.3 & p=0.1 & p=0 &\\ 
\hline
&&\multicolumn{6}{c|}{FB}&\\
sorting  & \textbf{503} & 3492 (784) & NA & NA  & NA & NA & NA&\\
maze  & 393 & \textbf{1.1e+4} & 1.6e+4 (414) & NA  & NA  & NA & NA&\\ 
&&\multicolumn{6}{c|}{FB+PU}&\\
sorting  & 103 & 104 & \textbf{160} & 342 (2) & 1295 (169) & 1287 (979)& NA&\\
maze   & 39 & 42 & 55  & 78 & \textbf{148} & 1349 (1)& NA&\\
\hline 
&&\multicolumn{6}{c|}{RL+FB}&RL\\
sorting& 503 & 1012 (58) & 1243 (231) & 1413 (574) & 1665 (759) & 2227 (758) & \textbf{2476 (716)}& 2476 (716)\\
maze  & 393 & 605  & 9904 & 1.4e+4 & 1.9e+4 & 2.4e+4 & \textbf{2.5e+4}&2.5e+4\\
&&\multicolumn{6}{c|}{RL+FB+PU}&RL+PU\\
sorting&  103  & 104  & 300 & 520 & 696 & 1069 & \textbf{1298}&1298\\
maze  &39  & 40  & 44  & 69  & 429 & 1278 & \textbf{3011} &3011\\
\hline 
\end{tabular}

\label{tab:fb-sparse} 
\end{table}

In the maze navigation task, the baseline model FB is robust down to a probability of feedback $Pr(fb|s,a) \geq 0.9$;  
while FB+PU is robust down to a probability of feedback $Pr(fb|s,a) \geq 0.3$. 
These results show that our framework improves the robustness of the learning process against feedback sparsity.
However, the robustness is still limited to a certain level of sparsity.

This limitation can be alleviated by using an additional evaluation source that would take over the learning process whenever feedback is lacking.
Table \ref{tab:fb-sparse} shows that when evaluative feedback is combined with a reward function, the learning process becomes robust against all levels of feedback sparsity.
Generally, we observe that the more sparse the feedback, the more the models tend to behave like when learning only from the reward function. 
So, in the extreme case with no feedback, the baseline RL+FB is equivalent to RL, and RL+FB+PU is equivalent to RL+PU.
In both cases, convergence is guaranteed by the reward function.
Note that for the object sorting task, the baseline model RL+FB does not completely converge to the 4-steps solution whenever feedback becomes sparse. 
However, it still converges to the near-optimal 5-steps solution; but here we only report the results for the 4-steps solution for comparison.
The RL+FB+PU model, on the other hand, is robust against all levels of feedback sparsity in both domains. 

Globally, we observe that for a given number of feedback, RL+FB+PU improves the convergence rate with respect to RL+FB. 
The convergence rate of RL+FB ranges from the performance of RL to the performance of FB with full feedback.
The RL+FB+PU model benefits from a larger range of convergence rates.
Without teaching signals, it performs like RL, while the maximum convergence rate goes beyond the performance of FB-with-full-feedback.

To sum up, we can say that our framework improves the robustness against feedback sparsity when learning only from evaluative feedback, but it is still limited to a certain level of feedback sparsity.
Using a reward function in addition to evaluative feedback makes the learning process robust against all levels of feedback sparsity.
But yet, for the same number of feedback, our framework improves the convergence rate, with respect to learning without unlabeled instructions.

\subsection{Erroneous feedback}
\label{fberr}
We now consider the condition where the teacher may provide erroneous feedback. 
Under this condition, instructions are correct and neither instructions nor feedback are sparse.
On every time-step, there is a probability that the feedback is not consistent with the teacher's policy.

When learning only from evaluative feedback, we observe the same behaviour in both domains (Table \ref{tab:fb-err}). 
Whether instructions are used or not, the learning process is not robust to a probability of error $Pr(fb_t \neq fb_T^*(s_t,a_t)) \geq 0.5$.
This is a predictable result, since evaluative feedback is binary.
For a probability of error $Pr(fb_t \neq fb_T^*(s_t,a_t)) = 0.5$, evaluative feedback provides no information about the policy.
Below this level of error, however, our model and the baseline completely converge in both domains.
Nevertheless, our model still outperforms the baseline for the same level of erroneous feedback.

When evaluative feedback is combined with a reward function, no significant improvement is observed.
However, this has to be put into perspective with the fact that feedback is not sparse.
We can expect more effect from the reward function with sparse feedback.
Nevertheless, these results show that, whether a reward function is used or not, our framework is robust against erroneous feedback and improves the convergence rate, if the teacher is optimal most of the time.

\begin{table}
\center
\caption{Model performance under erroneous feedback. The probability of receiving an erroneous feedback is $p = Pr(fb_t \neq fb_T^*(s_t,a_t))$. The reported numbers correspond to the 99th percentile of the number of steps until convergence. In parentheses, the number of sessions over 1000 that did not converge.}
\begin{tabular}{|l|cccc|}
\hline
& p=0 & p=0.1 & p=0.3 & p=0.5 \\ 
\hline
&\multicolumn{4}{c|}{FB}\\  
sorting  & 503 & 892 & \textbf{4065} & NA \\
maze  &393& 871 & \textbf{5681} & NA  \\ 
&\multicolumn{4}{c|}{FB+PU}\\  
sorting & 103 & 142  & \textbf{643} & NA  \\
maze  &39  & 47 & \textbf{135} & 4.6e+4 (997) \\
\hline 
&\multicolumn{4}{c|}{RL+FB}\\
sorting & 503  & \textbf{859} & 4184 (12) & NA \\
maze  & 393 & 729 &\textbf{5163} & NA \\
&\multicolumn{4}{c|}{RL+FB+PU}\\
sorting & 103  & 136 & \textbf{512} & 2.3e+5 (980) \\
maze  & 39  & 48 & \textbf{135} & 3.6e+5 (809) \\
\hline 
\end{tabular}

\label{tab:fb-err} 
\end{table}

\subsection{Interaction load}
\label{interaction-load}
To evaluate the cost of our method in terms of interaction load, we measure the total number of required teaching signals, as described in Section \ref{load-definition}.
We assess the cost of using unlabeled instructions in addition to evaluative feedback with respect to only using evaluative feedback.
We only consider the ideal case where all teaching signals are correct and not sparse.

\begin{table}
\begin{center}
\caption{Interaction load measured as the maximum number of required teaching signals over 1000 training sessions.}
\begin{tabular}{|l||l| l||l| l|}
\hline
&\multicolumn{2}{l||}{Object sorting}&\multicolumn{2}{l|}{Maze navigation}\\
\cline{2-5}
& FB+PU & FB & FB+PU & FB\\
\hline
\#feedback                  & \textbf{106} & 563 & \textbf{46} & 568\\
\#instructions              &    53        &\textbf{0}&   181    & \textbf{0}\\
\#total                  & \textbf{159} & 563 & \textbf{227} & 568\\
\hline
\end{tabular}
\end{center}
\label{load}
\end{table}

Table \ref{load} reports the maximum number of required teaching signals over 1000 training sessions for both domains (see Fig. 5 in supplementary material).
Without using instructions, the teacher needs to provide at most 563 feedback for the object sorting domain and 568 feedback for maze navigation.
With our framework, these numbers are reduced to 159 teaching signals for the object sorting domain ($81\%$ fewer feedback, $72\%$ fewer teaching signals) and 227 teaching signals for maze navigation ($91\%$ fewer feedback, $60\%$ fewer teaching signals).
This demonstrates that our framework reduces the number of evaluative feedback and the total number of required teaching signals.

\subsection{Summary}
The experimental results obtained in simulation can be summarized as follows:

\textbf{Ideal case:} When teaching signals are correct and not sparse, our framework improves the convergence rate with respect to learning without unlabeled instructions.

\textbf{Sparse instructions:} When learning from evaluative feedback, our framework is robust against all levels of instruction sparsity, and improves the convergence rate with respect to not using unlabeled instructions.
However, when learning from a reward function, the existence of multiple possible interpretations can prevent the learning process from converging.
This only happens in domains with multiple optimal policies and when instructions are below a certain level of sparsity.
When the reward function is combined with evaluative feedback, our framework becomes robust against all levels of instruction sparsity, as feedback enables the teacher to rectify misinterpreted instructions.

\textbf{Erroneous instructions:} Our framework is robust against erroneous instructions and improves the convergence rate, if the probability of receiving erroneous instructions is lower than $0.3$.

\textbf{Sparse feedback:} When learning only from evaluative feedback, our framework improves the convergence rate and the robustness of the learning process against feedback sparsity.
However, it is still limited to a certain level of sparsity.
With a reward function, the learning process becomes robust against all levels of feedback sparsity.

\textbf{Erroneous feedback:} Our framework is robust against erroneous feedback and improves the convergence rate as long as the teacher provides correct feedback most of the time.

\textbf{Interaction load:} In the ideal case, our framework reduces the number of evaluative feedback and the total number of required teaching signals.

\section{Experiment with a real robot}
\label{baxter}
In this section, we evaluate our framework with a real robot and a real human teacher on the object sorting task (Fig. \ref{fig:scenario-a}).
We assess the performance of the TICS architecture when using unlabeled instructions with respect to only using evaluative feedback.
For the sake of simplicity, we do not consider a predefined reward function.
In order to assess the scalability of our framework to different task complexities, we contrast two experimental conditions by varying the complexity of the state-space representation.

In this experiment, we use a slightly different implementation of the TICS architecture using Q-learning instead of Actor-Critic.
So, we first detail our methods in Section \ref{baxter-methods}.
Then, we present the experimental setup with the real robot in Section \ref{baxter-setup}.
The experimental results are reported in Section \ref{baxter-results}.
These results confirm that our framework improves the convergence rate of the learning process and reduces the number of required teaching signals, in line with the results reported in simulation.

\subsection{Methods}
\label{baxter-methods}

In this experiment, the TICS architecture is based on the Q-learning algorithm instead of Actor-Critic.
In the Task Model, evaluative feedback is converted into numerical values $r\in\{-1,1\}$ and used in a reward-shaping fashion for updating a Q-function, with $\gamma=0$ and $\alpha = 0.3$ \footnote{With myopic discounting ($\gamma=0$) \cite{knox_reinforcement_2012}, the Q-values play the same role as policy parameters in Actor-Critic.
So, this method is still compatible with our view about evaluative feedback as information about the policy.}.

For interpreting instructions, we use the Reward-based Updating method (RU) that considers instruction signals as an alternative state-space for learning a Q-function within the Instruction Model, in the same way as in the Task Model (cf. Section \ref{sec:IM}).
As shown in Section \ref{ideal}, RU has the same performance as PU, when learning from non sparse evaluative feedback and instructions.
So, we will refer to our model in this experiment as FB+RU.

We rely on the same implementation of the Contingency Model as in Section \ref{methods}. 
However, we implement a different shaping method that updates the Q-values of the Task Model towards the Q-values of the Instruction Model, and then uses the policy of the Task Model for decision-making.
We consider a greedy action-selection strategy.
As the learning process is completely guided by the teacher, random exploration is not required.

\subsection{Experimental setup}
\label{baxter-setup}

The experimental set-up (Fig. \ref{fig:scenario-a}) is composed of a Baxter Research Robot facing a table on top of which we place three magnets. 
The magnets allow placing objects at three different positions on the table: left, middle and right. 
Pictures of the objects placed at the middle position can be taken with a webcam placed between the robot and the table. 
A Microsoft Kinect\footnote{https://dev.windows.com/en-us/kinect, accessed 20-12-2014. We use a modified version of the Kinect V2 ROS client/server provided by the Personal Robotics Laboratory of Carnegie Mellon University. https://github.com/personalrobotics/, Last accessed 20-12-2014.} V2 sensor is placed on top of the robot's head and used for extracting feedback and instruction signals from the human teacher. 
The screen on the robot's head is employed as a transparency device to display the current task state, the associated instruction signal, the performed action and the perceived reward.
The robot is controlled by an implementation of the TICS architecture on ROS \cite{ros_2009}.

\subsubsection{Human teacher}
Throughout a training session, the human teacher\footnote{The first author of this paper.} uses either only feedback (FB) or feedback plus unlabeled instructions (FB+RU). 
Feedback is divided in two categories ${fb \in \{head\_nod, head\_shake\}}$. 
By convention, head nods are converted into a positive reward, while head shakes are converted into a negative reward. 

Instruction signals are defined over the teacher's hand gestures. 
For each hand, the system recognizes five states 
$h \in \{pointing\_right, pointing\_left,$ $pointing\_middle,\\ raised\_open, raised\_closed \}$, resulting in $35$ possible gestures using either one or both hands. 
In this work, we use only one instruction signal per action; so we only use nine gestures.  

In all sessions, the teacher provides feedback for every step (non sparse feedback). 
In sessions including unlabeled instructions, the teacher provides an instruction signal only if the robot did not receive any instruction for the current state or if the recorded signal is erroneous.

\subsubsection{Experimental conditions}

We follow the same protocol as \cite{suay_effect_2011} by considering two different conditions:

\textbf{Small state space:} In this condition, the object descriptor $D= (TYPE) ; \\ TYPE \in \{Plain, Pattern, Unknown\}$ contains one single variable based on the number of Speeded-Up Robust Features (SURF) \cite{bay_speeded-up_2008} descriptors of the object. The $Plain$, $Pattern$ and $Unknown$ values are obtained by thresholding the number of extracted SURF descriptors. If this number is less than 50, the object is considered as $Plain$. Otherwise it is considered as $Pattern$. 
The number of different task states resulting from this representation is 72.

\textbf{Large state space: } In this condition, the object descriptor $D= (TYPE,\\ COLOR, SIZE)$ contains three variables: $TYPE \in \{Plain, Pattern, Unknown\}$ describes the number of SURF descriptors as in the previous condition. $COLOR \in \{Red, Green, Blue, Unknown\}$ describes the dominant color of the object that can be red, green or blue. $SIZE \in \{Large, Small, Unknown\}$ describes the area of the bounding box of the object. This representation yields 864 task states.

In order to assess the scalability of our framework to different task complexities, we compare our model FB+RU to the baseline model FB in both the small and the large state space conditions. 
We conduct four training sessions with each model in each condition, which results in 16 sessions. 
In each training session, six different objects are presented one by one to the robot in a specific order. 
Four different orders were chosen randomly beforehand and the same orders were employed for both models in both conditions. 
Each session ends when all the six objects have been presented twice for the small state space condition and three times for the large state space condition, whether learning converged or not.
A video of one training session can be found online\footnote{\url{https://youtu.be/TK9SwFedtUc}}.

\subsection{Experimental results}
\label{baxter-results}

\begin{figure}
\centering
\subfloat[Small state space.]{\includegraphics[trim=1.5cm 0.6cm 1.7cm 2cm, clip=true,width=0.48\columnwidth]{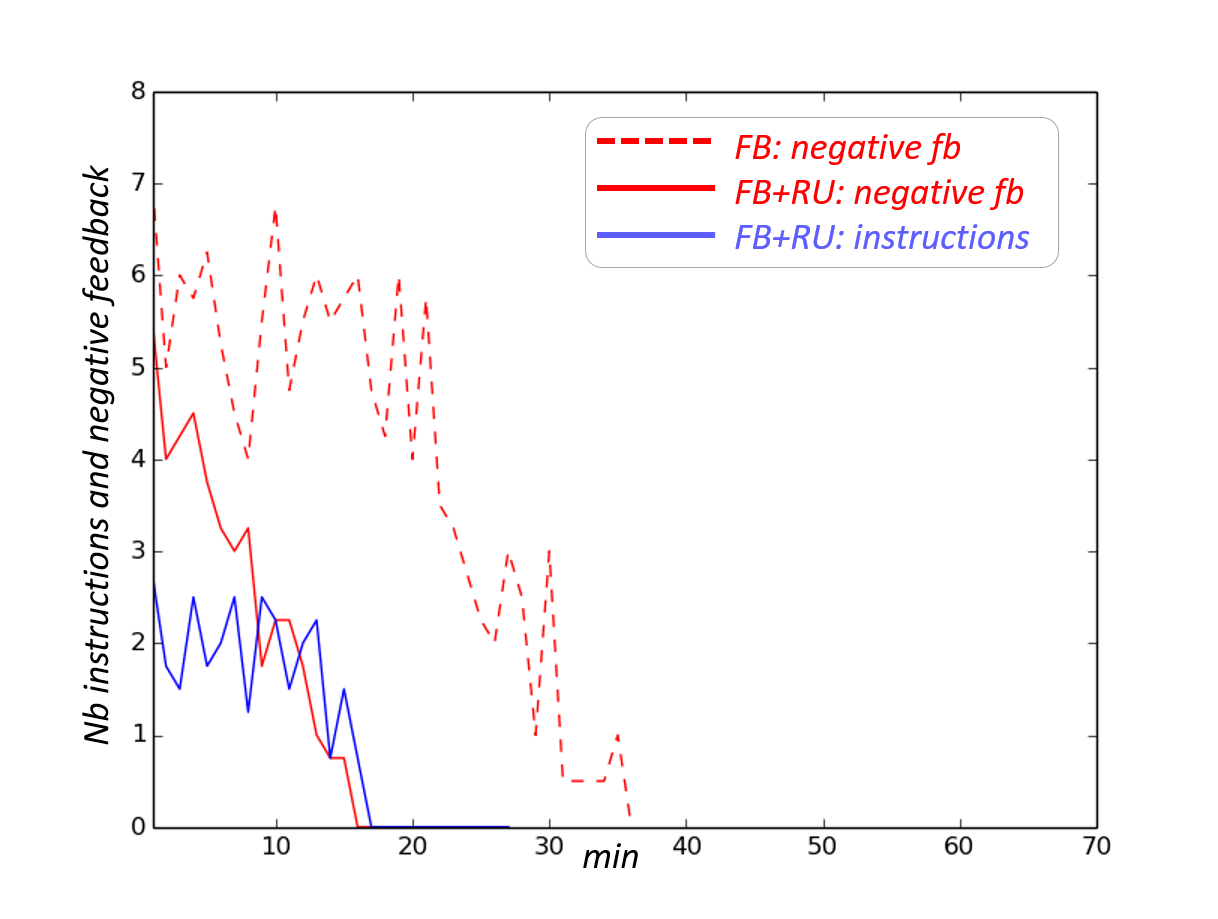}\label{fig:1a}}
\hfill     
\subfloat[Large state space.]{\includegraphics[trim=1.8cm 0.6cm 1.7cm 2cm, clip=true,width=0.48\columnwidth]{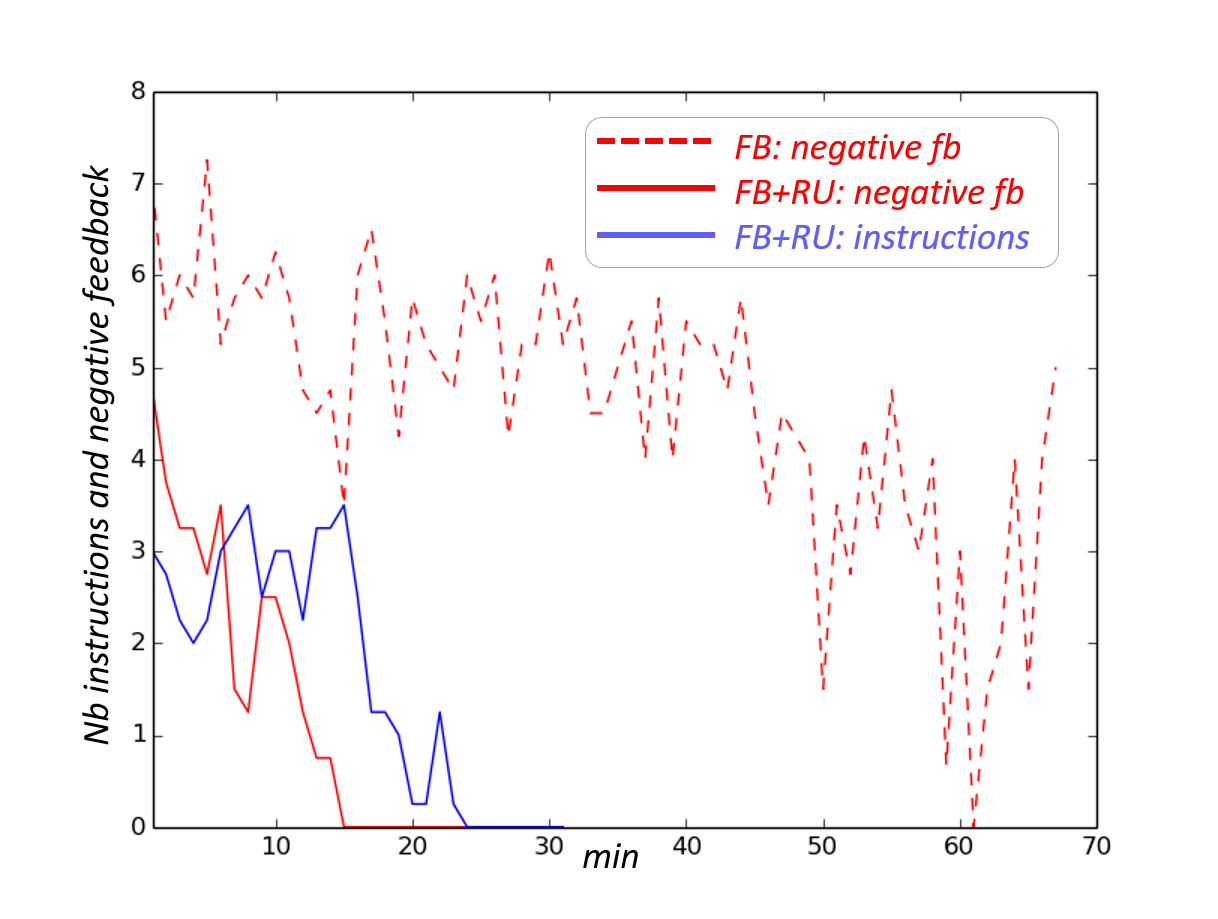}\label{fig:1b}}
\caption{Number of instructions (blue) and negative feedback (red) over time.}
\label{fig:exp} 
\end{figure}

Figure \ref{fig:exp} reports the evolution of the number of provided instructions and negative feedback over time for each condition. 
The results are averaged over the four sessions. 
We can see that in the small state space condition, the baseline model converges after at most $36$ minutes, while our model converges within $17$ minutes. 
In the large state space condition, the baseline model does not completely converge after an hour of training, while our model converges after at most $24$ minutes.

\begin{table}
\begin{center}
\begin{tabular}{|l||l| l||l| l|}
\hline
&\multicolumn{2}{l||}{Small state space}&\multicolumn{2}{l|}{Large state space}\\
\cline{2-5}
& FB+RU & FB & FB+RU & FB\\
\hline
training time (min)          & \textbf{25}  & 33  & \textbf{31}  & 67\\
\#steps                     & \textbf{135} & 235 & \textbf{166} & 470\\
\#feedback                  & \textbf{135} & 235 & \textbf{165} & 466\\
\#negative feedback         & \textbf{42}  & 135 & \textbf{33}  & 296\\
\#instructions              &    29        &\textbf{0}&   50    & \textbf{0}\\
\#states                    & \textbf{36} & 51  & \textbf{61}  & 108\\
\#undesired states          & \textbf{8}   & 14  & \textbf{8}   & 23\\
\#steps in undesired states & \textbf{10}  & 29  & \textbf{10}  & 70\\
\#Q-values                  & \textbf{265} & 164 & \textbf{450} & 359\\
\hline
\end{tabular}
\end{center}
\caption{Experiment statistics. The results are averaged over four training sessions. Training time, number of steps, number of provided feedback, negative feedback and instruction signals, number of explored states, number of undesired states, number of steps spent in undesired states and number of learned Q-values.}
\label{stats}
\end{table}

Table \ref{stats} reports various statistics over the training sessions such as the training time, the number of steps, number of teaching signals, number of explored states, number of undesired states and the number of steps spent in undesired states. 
Undesired states define situations in which the robot is holding the object with the wrong hand, or is holding the object while its descriptors are unknown (this may happen if the robot takes a picture after picking the object). 
We also report the size of the Q-function measured as the number of state-action pairs for which the algorithm learns a value.

The experimental results are consistent with those obtained in simulation and with the results reported by \cite{suay_effect_2011}. 
They show that our model reduces considerably the number of steps and training time ($42\%$ fewer steps for the small state condition and $64\%$ in the large state condition). 
It is also more efficient by achieving better performance with less interactions ($30\%$ less teaching signals for the small state condition and $53\%$ in the large state condition). 
The robot also explores fewer states (respectively $29\%$ and $43\%$ fewer in each condition), fewer undesired states (resp. $42\%$ and $65\%$ fewer) and spends less time in these states (resp. $65\%$ and $85\%$  less). 
This reflects a more efficient exploration strategy for our model.

Finally, when using only evaluative feedback, the robot learns in average $3$ action values per state, compared to $7$ action values per state when also using unlabeled instructions.  
This means that our model can determine more efficiently the Q-function in less time. 

This experiment validates the results we obtained in simulation.
It demonstrates that our framework reduces both the training time and the number of required teaching signals with respect to not using unlabeled instructions.

\section{Discussion}
\label{discussion}

Our experimental results, both in simulation and with a real robot, demonstrate the effectiveness of our framework in improving the convergence rate of the learning process and in reducing the total number of required teaching signals.
This performance can be explained by a reduction in the complexity of the learning process.

To simplify the complexity analysis, we only consider the ideal case where teaching signals are correct and non sparse.
Also, we do not consider the effect of the reward function, since it is negligible compared to the effect of evaluative feedback in the ideal case.
Without instructions, learning the task requires to derive a policy over the entire state space, which amounts to computing $|S| \times |A|$ state-action preference values.
With our framework, task learning is divided into two processes: interpreting instructions and shaping.
Interpreting instructions requires to derive a policy over instruction signals, which amounts to computing $|I| \times |A|$ signal-action preference values.
Shaping requires to use the information about the optimal action in every task state.
This is done by associating an instruction signal to each different task state, which requires $|S|$ operations.
So, with our framework, the learning process requires $|S|+|I| \times |A|$ operations instead of $|S| \times |A|$.
Our framework reduces the complexity of the learning process only if ${|I| < |S| \times (1-\frac{1}{|A|})}$, which is equivalent to $ |I|<|S|$ as $|A|$ gets large.

The number of provided instructions $|I|$ can be written as $|I|=|O| \times |C|$, where $|O|$ is the number of different optimal actions required for executing the task, and $|C|$ is the number of different instruction signals used by the teacher for indicating each action.
So, the more task states sharing the same optimal action there are, and the fewer different signals are used by the teacher per action, the better the gain of our method.
In our experiments, we considered domains such that $|O|=|A|$ and $|A|<|S|$; and we assumed $|C|=1$. 
So, the condition $|I|<|S|$ was satisfied.

The reduction in complexity resulting from our framework can be explained by the role played by unlabeled instructions, which serve as a bottleneck that shifts the complexity of the learning process from the task state space to the signal state space.
This scheme can be considered as a hybrid learning method that lies between reinforcement learning and supervised learning and combines their benefits (Fig. \ref{fig:sl-rm}).
In a pure RL scheme, learning is mainly based on the exploration of the state-action space and suffers from slow convergence.
Supervised learning, on the other hand, is more straightforward but requires predefined labels.
In our framework, instruction signals constitute labels whose meanings are learned by RL and used for learning the task in a supervised fashion.
More precisely, the Policy-based Updating method (PU) interprets instructions by using the TD error of the task-learning process; and shaping mainly depends on the Contingency Model which associates instruction signals to task states in a supervised learning way.
The PU method proposed in this paper alleviates the limitation of the standard interpretation method, RU, which only works with non sparse instructions.

Our work presents similarities with autonomous reward-shaping methods \cite{konidaris_estimating_2004,konidaris_autonomous_2006,marthi_automatic_2007,grzes_online_2010}, that share the common idea of learning a value function on an alternative state-space, and then using this function for shaping the task in the original state-space. 
Our framework is based on a similar idea, that instruction signals define an alternative state-space in which a policy is learned, and used for shaping the policy of the task state-space. 
First, as with external sensors in \cite{konidaris_autonomous_2006}, instruction signals represent additional sensory readings that are not part of the ``problem-space".
Second, as in \cite{grzes_online_2010}, instruction signals can be considered as a clustering of the ``grond-space", where each signal regroups all states in which it has been provided.
The main difference is that with our framework, the similarity measure defining the clustering is not based on the topographic proximity of task states, but rather on whether or not these states share the same optimal action.

\begin{figure}
\begin{center}
\includegraphics[scale=0.45]{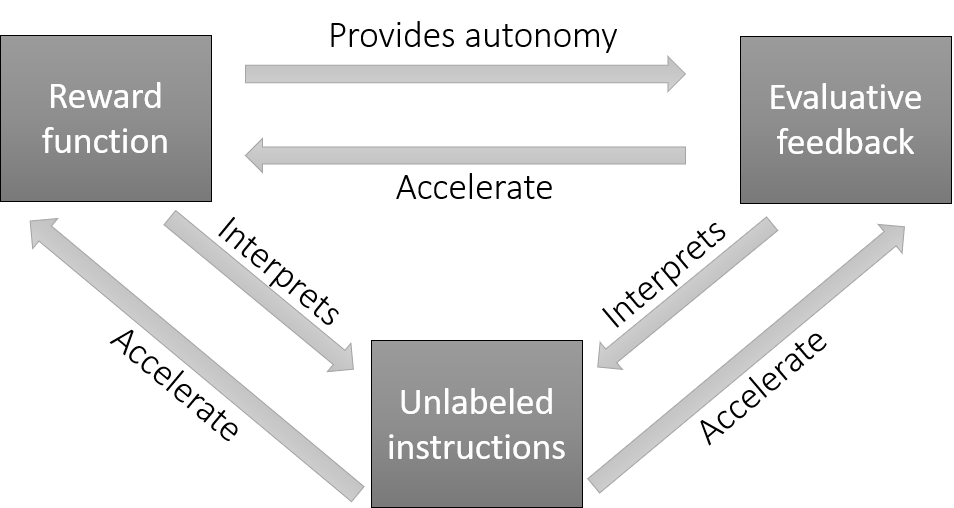}
\caption{Relationship between different information sources in our framework.}
\label{fig:inf-sources}
\end{center}
\end{figure}

In addition to reducing the complexity of the learning process, our framework provides a flexible way for switching between different learning modalities: learning from a reward function,  learning from evaluative feedback and learning from instructions.
This lets the teacher benefit from the advantages of each source of information (Fig. \ref{fig:inf-sources}).
Evaluative feedback accelerates the learning process with respect to the reward function.
Instructions accelerate the learning process with respect to both evaluative feedback and the reward function.
Finally, the reward function provides autonomy with respect to evaluative feedback, in both task learning and interpreting instructions.

Despite all these advantages, our framework still suffers from several limitations. 
For instance, we assumed that the robot was able to detect a predefined set of instruction signals, which requires a prior segmentation of these signals.
This constitutes a limitation for our system, because even with an important category of discrete signals, the teacher is still constrained by the signals that the system is able to detect.
On possible solution to this limitation would be to perform an online segmentation of the teacher's gestures.

In this paper, we also assumed that the teacher had a single preferred action for every task state, and only one signal is used per action.
The main motivation for doing so is to handle contradictory and erroneous instructions.
In fact, we assumed that the existence of multiple instructions was synonymous to contradiction.
So, at every time step, only the most likely instruction signal is extracted from the Contingency Model for interpretation and shaping.
However, this implementation limits the possibility for the teacher to indicate the same action with different signals or to suggest several actions for the same state, which can be useful in problems with multiple optimal policies.
A possible extension would be to consider all the signals that have been detected in a given state, and to use their contingencies as weights for both interpretation and shaping.

Another limitation resides in the transparency-based teaching protocol, in which the teacher has access to the signal that is associated to the current state. 
The main motivation behind this protocol is to limit the number of interactions with the robot, so the teacher needs to provide instructions only when required.
Without such a protocol, our method would be less effective in reducing the number of teaching signals, but would still be as effective in accelerating the learning process.
Nevertheless, this protocol requires a transparency device and is not easy to implement in all real robots. 
In our case, we used the screen of the Baxter robot, but other solutions such as intentional actions \cite{knox_learning_2012} or gazing behaviours \cite{thomaz_transparency_2006} can also be employed.
There are some other alternatives such as active learning protocols where the robot asks explicitly for teaching signals when required. 
However this solution also has its limitations. 
For example, it can be constraining and annoying for the teacher to not control the pace of the interaction. 
All these aspects related to the teaching protocol are very important and require further investigations in future work.
More extended user studies are necessary to provide external validation of the assumptions about how people would be interacting with our system.

\section{Conclusion}
\label{conclusion}

This paper presents a novel framework for interactively shaping a robot behaviour with unlabeled human instructions.
The key idea is to reduce the complexity of a task-learning process through unlabeled instruction signals.
These signals are interpreted by the robot, and used simultaneously for accelerating the task-learning process.

This approach has several advantages. 
First, using unlabeled instructions offers more adaptability to the preferences of the teacher, by providing more flexibility in the choice of signals and in their meaning. 
Second, this reduces the required engineering effort, by removing the constraint about encoding the meaning of each signal.

We implemented our framework as a modular architecture (TICS) based on four components: a Task Model, an Instruction Model, a Contingency Model and a Shaping Component.
This modular architecture makes it possible to integrate several sources of information, such as a reward function, evaluative feedback and instructions.
This enables the teacher to switch between different teaching modalities in order to benefit from each source of information.
Although in this paper we proposed one particular implementation, the modularity of our architecture makes it possible to imagine several extensions for each component.
For instance, different algorithms can be designed for task learning, for computing contingency, for interpreting instructions and for shaping.

Finally, the experimental results reported in this paper demonstrate the effectiveness of our framework in accelerating the task-learning process and in reducing the number of required teaching signals.
The complexity reduction performed by our framework provides a novel perspective for combining Reinforcement Learning and Supervised Learning paradigms.

\section*{Supplementary material}
\label{suppl}
Supplementary material includes five figures and can be found with this article online.

\bibliographystyle{abbrv}

\end{document}